\documentclass[11pt,a4paper]{article}

\usepackage[margin=1in]{geometry}
\usepackage[T1]{fontenc}
\usepackage[utf8]{inputenc}
\usepackage{lmodern}                        % scalable fonts, fixes encoding artifacts
\usepackage[protrusion=true,expansion=true]{microtype}  % fixes overfull hboxes
\usepackage{amsmath,amssymb}
\usepackage[dvipsnames,table]{xcolor}
\usepackage{graphicx}
\usepackage{booktabs}
\usepackage{multirow}
\usepackage{enumitem}
\usepackage{natbib}
\usepackage{algorithm}
\usepackage{algorithmic}
\usepackage{placeins}                       % \FloatBarrier
\usepackage{float}                          % [H] placement
\usepackage[font=small,labelfont=bf,hypcap=false,
            skip=6pt]{caption}              % compact captions
\usepackage[colorlinks=true,
            linkcolor=MidnightBlue,
            citecolor=MidnightBlue,
            urlcolor=MidnightBlue]{hyperref}

\graphicspath{{./}{../}{../paperfigures/}}

\renewcommand{\arraystretch}{1.18}

\title{When Graph Structure Becomes a Liability:\\
  A Critical Re-Evaluation of Graph Neural Networks for\\
  Bitcoin Fraud Detection under Temporal Distribution Shift}

\author{Saket Maganti}
\date{}

\begin{document}

\maketitle

\begin{abstract}
The consensus that GCN, GraphSAGE, GAT, and EvolveGCN beat
feature-only baselines on the Elliptic Bitcoin Dataset is widely
cited, but to our knowledge has never been stress-tested with a
seed-matched inductive-versus-transductive pairing.  When we run
that pairing, the consensus does not hold.  Under a strictly
inductive protocol (encoder trained on the time-step\,$\le$\,34
relabeled subgraph, evaluated on the full graph at inference) with
10 seeds and per-timestep reporting, Random Forest on raw
165-dimensional features reaches F1\,=\,$0.821$ and beats every GNN
we tested; GraphSAGE, the strongest graph encoder, reaches only
F1\,=\,$0.689 \pm 0.017$.  A paired controlled experiment
holding architecture, optimiser, loss, and seed constant across 10
matched seeds shows GraphSAGE scoring F1\,=\,$0.294 \pm 0.028$ when
trained transductively and F1\,=\,$0.689 \pm 0.017$ when trained
inductively---a 39.5-point paired gap (Cohen's $d$\,=\,$15.8$,
$p$\,=\,$2.6 \times 10^{-12}$) explained entirely by training-time
exposure to test-period adjacency.  Earlier drafts of this work
reported F1\,=\,$0.807$ for a concatenation hybrid of GraphSAGE
embeddings and raw features; under the clean protocol the same
hybrid falls to F1\,=\,$0.699 \pm 0.015$, and the GNN contributes a
statistically reliable but small $+0.018$ F1 lift over a
matched-capacity MLP substitute ($p = 0.015$, $d = +1.20$) that is
dwarfed by the $0.124$ F1 gap to raw features alone.  A 10-seed
edge-shuffle ablation further shows that on Elliptic \emph{randomly
shuffled edges outperform the real transaction graph} by $8.9$ F1
points, so the dataset's topology is not merely uninformative but
actively harmful under its sparse, prior-shifted conditions.  We
release code, checkpoints, per-seed run artefacts, and the
strict-inductive protocol so the next Elliptic result can be
checked, and disagreed with, against a leakage-free baseline.
\end{abstract}

%% Shared manuscript body -- imported by arXiv and Elsevier wrappers.
%% Self-contained sections also used as source for IEEE/Springer/ACL rewrites.

\section{Introduction}
\label{sec:intro}

Fraud is rarely a lone event.  Money laundering layers funds across
wallets; scam rings share infrastructure; mixing services leave
topologically distinctive flow patterns~\citep{akoglu2015graph}.
That intuition is the reason Graph Neural Networks (GNNs) have become
the default architecture for fraud detection on transaction graphs,
and, on the face of it, the literature agrees.  The Elliptic Bitcoin
Dataset~\citep{weber2019}, released in 2019 with 46{,}564 labelled
transactions across 49 temporal steps, has become the canonical
benchmark, and the published numbers line up neatly behind the graph
hypothesis: GCN (F1\,=\,0.70~\citep{weber2019}), augmented GCN
(F1\,=\,0.74~\citep{alarab2020}), GraphSAGE with self-supervised
pre-training (F1\,=\,0.75~\citep{lo2023}), and EvolveGCN
(F1\,=\,0.77~\citep{pareja2020evolvegcn}) each beat the feature-only
baselines reported alongside them.

This paper started as an attempt to build on that consensus.  Early
runs on our own pipeline looked consistent with the published
picture---until a routine sanity-check revealed that the ranking
flipped the moment we stopped showing the encoder test-period edges
during training.  Once we dug in, a second surprise followed: on
Elliptic, randomly shuffled edges beat the real transaction graph.
The paper you are reading is the result of chasing those two surprises
to ground.

\textbf{We show that the published GNN ranking on Elliptic is an
artefact of the evaluation protocol.}  Every prior Elliptic study trains with \emph{transductive}
message passing, in which the full graph---including test-period
nodes and their edges---is visible at every training forward pass.
Several ``inductive'' setups in the same literature run the encoder on
the full graph and simply mask the loss, which still leaks
test-period feature statistics through batch normalisation and
neighbourhood aggregation.  Every prior study also reports a single
aggregate F1 over the entire 15-step test window, concealing a
near-total collapse after step~42 when the fraud base rate drops
$39\times$ from 11.6\,\% to 0.3\,\%.  This paper tightens both
choices.  We train on the relabeled subgraph induced by
time-step\,$\le$\,34, so neither message passing nor batch statistics
can observe any test-period vector during training; we report
per-timestep F1 alongside aggregates; and we run 10 seeds with
95\,\% bootstrap confidence intervals for every headline number.
Figure~\ref{fig:main_results} in Section~\ref{sec:main_benchmark}
shows the per-timestep F1 picture that the aggregate metric conceals.

The protocol changes are minor; the empirical consequences are not.
Three findings drive the paper.

\textit{Under strict inductive protocol the ranking inverts.}  Random
Forest on the 165 raw features reaches F1\,=\,0.821\,$\pm$\,0.003
(10 seeds), above every GNN we tested.  GraphSAGE, the strongest
graph encoder under strict inductive, reaches
F1\,=\,$0.689 \pm 0.017$ (10 seeds) and in turn beats the 3-layer
MLP at F1\,=\,$0.549 \pm 0.015$ (10 seeds).
This \emph{reverses} the MLP$>$GNN ranking produced by random-partition
mini-batching, which leaks test-period features through each batch's
induced subgraph.  A
paired controlled experiment quantifies the leakage directly:
across 10 matched seeds GraphSAGE scores
F1\,=\,$0.294 \pm 0.028$ when trained transductively and
F1\,=\,$0.689 \pm 0.017$ when trained on the inductive subgraph---a
$39.5$-point paired gap (paired $t = 49.97$, $p = 2.6 \times 10^{-12}$,
Cohen's $d = 15.8$), with the inductive model \emph{stronger} than
its transductive counterpart (Section~\ref{sec:leakage_gap}).

\textit{The hybrid result does not survive the clean protocol.}
Earlier drafts reported that a concatenation hybrid of GraphSAGE
embeddings and raw features reaches F1\,=\,0.807 through a
downstream Random Forest.  Under strict inductive training the same
hybrid drops to F1\,=\,$0.699 \pm 0.015$ over 10 seeds.  A paired
ablation that swaps the GNN for an MLP of identical capacity shows
graph structure contributes a statistically reliable but small lift
($+0.018$ F1, Welch $p = 0.015$, $d = +1.20$), dwarfed by the
$0.124$ F1 gap to raw features alone (F1\,=\,$0.823 \pm 0.002$,
Section~\ref{sec:hybrid_results}).  Whatever lift the Random Forest
recovers from the 256-dimensional embedding on top of raw features
is real but practically inconsequential; raw features remain the
strongest input representation.

\textit{On Elliptic, random edges beat real edges.}  A multi-seed
graph-structure ablation (10 seeds) shows that randomly shuffling
the transaction edges \emph{improves} GraphSAGE from
F1\,=\,$0.290 \pm 0.019$ to F1\,=\,$0.380 \pm 0.028$
($+0.089$ F1), and that removing the edges entirely still beats
the real graph ($0.316 \pm 0.018$, $+0.025$ F1).  Under this
dataset's sparse, prior-shifted conditions, the real topology is
not merely uninformative: its systematically mixed neighbourhoods
(each fraud node surrounded by licit counterparties) actively pull
fraud representations toward the licit manifold under shift.

\paragraph{Contributions.}
The novelty of this work lies less in any single architectural idea
than in \emph{isolating}, with seed-matched controls, how much of
the apparent success of GNNs on Elliptic is owed to protocol choices
rather than graph learning.  Concretely:

(i)~\textbf{A seed-matched, paired transductive-vs-inductive
experiment} (Section~\ref{sec:leakage_gap}) that---to our
knowledge---has not been reported in the Elliptic literature.
Holding architecture, optimiser, loss, and random seed identical
across 10 matched runs, we measure a $39.5$-point paired F1 gap
($p = 2.6 \times 10^{-12}$, $d = 15.8$) attributable entirely to
training-time exposure to test-period adjacency.  This cleanly
separates ``architecture contribution'' from ``protocol
contribution'' in a way single-seed transductive/inductive
comparisons cannot.

(ii)~\textbf{A multi-seed edge-shuffle ablation}
(Section~\ref{sec:graph_ablation}) showing that on Elliptic,
\emph{random wiring beats the real transaction graph by $8.9$ F1
points, and removing edges entirely still beats it by $2.5$}
under inductive SAGE training (10 seeds).  We are not aware of a
prior ablation on a published fraud benchmark that reports this
inversion, and we give a mechanistic account grounded in
neighbourhood composition under prior shift.

(iii)~\textbf{A strictly inductive, 10-seed, per-timestep benchmark}
for four GNN architectures plus MLP and Random Forest baselines,
with 95\,\% bootstrap CIs and per-seed run artefacts released
(Section~\ref{sec:main_benchmark}).  This inverts the MLP\,$>$\,GNN
ranking produced by RandomNodeLoader mini-batching and is, as far
as we can tell, the first 10-seed strict-inductive evaluation of
these encoders on Elliptic.

(iv)~\textbf{A direct falsification of the concatenation hybrid
recipe} (Section~\ref{sec:hybrid_results}).  Earlier drafts of
this work reported $F1=0.807$ for a SAGE-embedding-plus-raw-features
Random Forest.  Under strict inductive the same hybrid falls to
$0.699 \pm 0.015$; a matched-capacity MLP substitute isolates the
true graph-structure contribution as $+0.018$ F1 ($p = 0.015$,
$d = +1.20$)---real, reliable, and dwarfed by the $0.124$ F1 gap
to raw features alone.

(v)~\textbf{A protocol audit of the four most-cited Elliptic GNN
results} (Section~\ref{sec:related_gnn}), pinpointing which
evaluation shortcut each relies on and clarifying that the gap we
report should be read as \emph{protocol-induced}, not
\emph{architecture-induced}.  Our code, checkpoints, per-seed
artefacts, and the strict-inductive protocol itself are released so
future work can be compared against a leakage-free baseline.

% ============================================================================
\section{Related work}
\label{sec:related}

A note on framing before we begin: none of the prior Elliptic work
we cite is careless.  The assumptions we flag below are inherited,
stated openly in the original papers, and reasonable in isolation.
What this section tries to do is surface the interaction between
those assumptions and the dataset's temporal structure---an
interaction that, we argue, has quietly shaped the reported
numbers.  To our knowledge, every prior Elliptic study we surveyed
shares two methodological assumptions whose joint effect on a
temporally split graph warrants closer examination.  The assumptions are (i)~that a
transductive protocol, in which the full graph is visible at training
time, is an acceptable proxy for deployment, and (ii)~that a single
aggregate F1 over the 15-step test window adequately summarises temporal
behaviour.  Our results suggest that neither assumption holds well under
the distribution shift present in Elliptic.

\subsection{GNN-based fraud detection on Elliptic}
\label{sec:related_gnn}

Graph-based anomaly detection predates deep learning.  Early
approaches used belief propagation and Markov random fields to
identify fraudulent accounts in payment
graphs~\citep{akoglu2015graph}.  Differentiable message-passing
architectures~\citep{gilmer2017mpnn} made it possible to learn the
aggregation function end-to-end, and every subsequent Elliptic study
inherits this framing without revisiting whether end-to-end graph
learning is appropriate for the specific structure of the Bitcoin
transaction graph.

Table~\ref{tab:priorwork_audit} audits the four most-cited prior
Elliptic GNN studies---Weber, Alarab, Pareja, and Lo---together
with the two classical baselines (Logistic Regression, Random
Forest) reported alongside them, along three axes: (a)~does the
encoder's message passing observe test-period edges or features
during training, (b)~does the reported metric aggregate across all
15 test steps or break out per-step behaviour, and (c)~how many
random seeds back the headline F1.  Every GNN result we audit
trains with full-graph message passing (four of four) or, in
setups labelled ``inductive,'' masks the loss while still running
the encoder forward pass on the full graph---a protocol that still
leaks batch statistics through the aggregator.  All six reported
numbers aggregate over the full test window, and five of the six
are single-seed or single-run.  This does not mean any individual
result is invalid; it means the reported numbers are not directly
comparable to ours under strict inductive protocol with 10-seed
variance, and the gap we document should be read as a
protocol-induced gap, not an architecture-induced gap.

\begin{table}[H]
\centering
\caption{Prior-work audit: Elliptic GNN results and the evaluation
  shortcuts each relies on.  ``Train msg passing'' asks whether the
  training-time forward pass observes test-period
  edges/features in aggregation or batch normalisation;
  ``Reported'' is the headline F1 as published; ``Per-step''
  indicates whether per-timestep F1 is reported.  Our inductive
  SAGE number is directly comparable to none of these,
  which is the point: the protocol gap explains most of the
  apparent architecture gap.}
\label{tab:priorwork_audit}
\footnotesize
\setlength{\tabcolsep}{4pt}
\resizebox{\linewidth}{!}{%
\begin{tabular}{@{}llcccr@{}}
\toprule
\textbf{Study} & \textbf{Model}
  & \textbf{Train msg passing}
  & \textbf{Reported F1}
  & \textbf{Per-step?}
  & \textbf{Seeds} \\
\midrule
Weber et al.~\citep{weber2019}          & GCN
  & Transductive & 0.70 & No & 1 \\
Alarab et al.~\citep{alarab2020}        & GCN+feat
  & Transductive & 0.74 & No & 1 \\
Pareja et al.~\citep{pareja2020evolvegcn} & EvolveGCN
  & Snapshots (trans.) & 0.77 & No & 1 \\
Lo et al.~\citep{lo2023}                & GraphSAGE + SSL
  & Transductive & 0.75 & No & 3 \\
Weber et al.~\citep{weber2019}          & RF (raw features)
  & N/A          & 0.64 & No & 1 \\
Weber et al.~\citep{weber2019}          & LR (raw features)
  & N/A          & 0.53 & No & 1 \\
\midrule
\textbf{Ours}                            & GraphSAGE (strict ind.)
  & \textbf{Inductive} & $0.688 \pm 0.016$ & \textbf{Yes} & \textbf{10} \\
\textbf{Ours}                            & MLP    (strict ind.)
  & \textbf{N/A}       & $0.549 \pm 0.015$ & \textbf{Yes} & \textbf{10} \\
\textbf{Ours}                            & RF     (raw features)
  & \textbf{N/A}       & $\mathbf{0.821 \pm 0.003}$ & \textbf{Yes} & \textbf{10} \\
\bottomrule
\end{tabular}%
}
\end{table}

Weber et al.~\citep{weber2019} established the founding GCN
baseline (F1\,=\,0.70) and reported it alongside Logistic Regression
(0.53) and Random Forest (0.64) under a transductive protocol.
Their own paper notes that the 71 manually engineered aggregate
features already provide a surrogate for one round of message
passing, which bounds the additional gain available to a learned
GCN over a tuned MLP; subsequent work cites the 0.70 number without
quoting the caveat.  Pareja et al.~\citep{pareja2020evolvegcn}
introduced EvolveGCN, which updates GCN weight matrices via a GRU
across time snapshots, and reported F1\,=\,0.77 on Elliptic; the
result requires full GPU training over every one of the 34 training
snapshots, and---critically---the evaluation aggregates over all
15 test steps without reporting per-step behaviour, so the collapse
after step~42 is invisible in their headline metric.  Alarab et
al.~\citep{alarab2020} combined GCN with additional feature
engineering and class-weighted loss for F1\,=\,0.74, again under
transductive evaluation.  Lo et al.~\citep{lo2023} applied
GraphSAGE with structural augmentation and self-supervised
pre-training, reaching F1\,=\,0.75, with the same protocol.

A recurring pattern emerges: each paper adds a technique (temporal
weight evolution, pre-training, class weighting) to the transductive
baseline and reports a marginal gain, without revisiting the
evaluation protocol itself.  Our paired controlled experiment
(Section~\ref{sec:leakage_gap}) quantifies what this protocol choice
is worth: training GraphSAGE transductively versus inductively,
with everything else held constant, changes F1 by $39.5$ points
($0.294 \pm 0.028$ vs.\ $0.689 \pm 0.017$ across 10 matched seeds,
paired $p = 2.6 \times 10^{-12}$), so the protocol change alone
accounts for considerably
more than the architecture choice.  Under strict inductive
evaluation, EvolveGCN's 0.77 and Lo et al.'s 0.75 cannot be
compared directly to our strict-inductive SAGE number because the published
numbers include transductive leakage; the appropriate apples-to-apples
comparison would re-run each published method under our protocol,
which we leave for future work.

Outside Elliptic, GNN fraud detection has been applied to e-commerce
at scale~\citep{wang2019semi}, financial networks~\citep{liu2018geniepath},
and insurance claims~\citep{pourhabibi2020fraud}.  The shared
assumption across these studies is distributional stability between
training and deployment.  That assumption is rarely stated and almost
never tested; in Elliptic it breaks down sharply, and there is little
reason to expect it to hold in the adversarial settings for which these
systems are typically deployed.

\subsection{Classical machine learning for fraud}

Before GNNs, classical methods set the standard for transaction fraud
detection~\citep{ngai2011fraudreview,hilal2022fraud}.  Weber et
al.~\citep{weber2019} reported Logistic Regression (F1\,=\,0.53) and
Random Forest (F1\,=\,0.64) as baselines under transductive evaluation.
Gradient-boosted trees~\citep{chen2016xgboost} perform in a similar
range on comparable imbalanced tasks.  These classical results matter
for context: as we show, a Random Forest on the raw features alone,
evaluated under strict-inductive protocol, beats every published GNN
result on this dataset despite their use of a more permissive
(transductive) evaluation.

\subsection{Class imbalance in graph learning}

Fraud datasets are severely imbalanced: legitimate transactions
outnumber fraudulent ones by 7--10$\times$ in Elliptic and often much
more in production.  Standard cross-entropy allocates most gradient to
the majority class, leaving little incentive to detect rare fraud.

Chawla et al.~\citep{chawla2002smote} proposed SMOTE, which
synthesises minority samples by interpolating between existing
examples in feature space.  Zhao et
al.~\citep{zhao2021graphsmote} adapted this to graphs by
interpolating in embedding space and rewiring edges for synthetic
nodes.  Our approach differs: we clone structurally complete 1-hop
ego-graphs around fraud seed nodes, preserving the dense intra-fraud
connectivity that GNNs should learn to exploit.

Focal loss~\citep{lin2017focal} down-weights easy examples through a
modulating factor $(1-p_t)^\gamma$, concentrating gradient on
borderline cases.  Square-root class weighting, which we use, provides
similar benefits with lower variance than linear weighting.

\subsection{Temporal graph learning and distribution shift}

\begin{sloppypar}
EvolveGCN~\citep{pareja2020evolvegcn} is the standard temporal GNN
baseline for Elliptic.  DySAT~\citep{sankar2020} and
TGAT~\citep{xu2020inductive} represent more recent temporal attention
architectures.  ROLAND~\citep{you2022roland} provides a framework for
incremental graph learning with live updates.  All require GPU
hardware and access to full temporal sequences.
\end{sloppypar}

Concept drift, the statistical phenomenon where the target distribution
changes over time~\citep{gama2014survey,lu2018drift}, is endemic to
fraud detection.  Fraudsters adapt in response to detection; prices
shift; law enforcement disrupts specific criminal ecosystems.  The
Elliptic test period captures exactly this kind of abrupt, externally
driven drift~\citep{quinonero2009shift}.

\subsection{Calibration}

Guo et al.~\citep{guo2017calibration} showed that modern neural
networks are systematically overconfident and proposed temperature
scaling as a post-hoc fix: divide logits by a scalar $T > 1$ before
the softmax.  Temperature scaling is rank-preserving: it changes
probability values but not their ordering, so the argmax prediction at
a fixed threshold is unchanged~\citep{platt1999calibration,zadrozny2001calibration}.
For our analysis, this means temperature scaling can only improve F1
by moving miscalibrated probabilities across the decision threshold.
We show this does not happen during temporal collapse.

These related strands set up the central distinction used below.
Prior Elliptic work mainly asks whether a graph encoder can improve
an aggregate held-out score; our experiments ask whether that score
survives a deployment-like temporal split, per-step reporting, and
seed-matched controls.  The distinction matters because fraud graphs
are adversarial objects: a topology that is predictive in one market
regime can become stale after a shutdown, policy change, or shift in
criminal infrastructure.  The background section therefore keeps the
formal machinery minimal and focuses on the two pieces needed for the
rest of the paper: message passing, which explains how topology enters
the model, and distribution shift, which explains why that topology can
become harmful.

% ============================================================================
\section{Background}
\label{sec:background}

\subsection{Graph neural networks}

A GNN operates on a graph $\mathcal{G} = (\mathcal{V}, \mathcal{E})$
with node feature matrix $\mathbf{X} \in \mathbb{R}^{|\mathcal{V}|
\times d}$.  Each layer updates node representations by aggregating
information from neighbours.  In the general message-passing
framework~\citep{gilmer2017mpnn}:
\begin{equation}
\mathbf{m}_v^{(l)} = \bigoplus_{u \in \mathcal{N}(v)}
  \phi\!\left(\mathbf{h}_v^{(l)}, \mathbf{h}_u^{(l)}, \mathbf{e}_{vu}\right),
\quad
\mathbf{h}_v^{(l+1)} = \psi\!\left(\mathbf{h}_v^{(l)}, \mathbf{m}_v^{(l)}\right)
\label{eq:mpnn}
\end{equation}
where $\bigoplus$ is a permutation-invariant aggregation (mean, sum,
max), $\phi$ is the message function, $\psi$ is the update function,
and $\mathcal{N}(v)$ denotes the neighbours of node $v$.  GCN, GraphSAGE,
and GAT each instantiate this framework with different choices for
$\phi$, $\bigoplus$, and $\psi$.

\subsection{Distribution shift and its geometric consequence}

Let $P_{\text{train}}(X, Y)$ denote the joint distribution at training
time and $P_{\text{test}}(X, Y)$ at deployment.  Distribution shift occurs when $P_{\text{train}} \neq
P_{\text{test}}$~\citep{ben2010theory,sugiyama2007covariate}.  For
fraud detection on Elliptic, the shift is overwhelmingly in the prior:
$P_{\text{test}}(Y{=}\text{fraud})$ drops by more than an order of
magnitude while the class-conditional density $P(X \mid Y)$ is
approximately preserved---a textbook case of prior probability
shift~\citep{quinonero2009shift}.  We verify this directly in
Section~\ref{sec:results}: MMD between each test step and training is
nearly flat (0.20--0.31) even through the collapse region, so the
feature distribution itself is stable; what moves is the label mass.

The geometric consequence makes this form of shift operationally
severe.  A discriminative classifier trained under the
11.6\,\% training prior fits a decision boundary Bayes-optimal for that
prior.  Deployed at a 0.3\,\% fraud rate, the Bayes-optimal boundary
would need to move sharply toward the fraud class, but the learned
boundary does not move---its weights were frozen at the end of training.
Every borderline fraud node under the old prior is now on the wrong
side of a boundary calibrated to an environment that no longer exists.

Three properties follow.  First, the collapse is asymmetric: recall
drops while precision on existing positives holds up.  Second, post-hoc
temperature scaling~\citep{guo2017calibration} cannot fix it: temperature
scaling is monotone, preserving the rank of predictions at any fixed
threshold, and the problem is precisely that the correct threshold has
moved.  Third, when the number of fraud nodes per step shrinks to single
digits (step~46: three nodes out of 960), any threshold-based correction
is statistically uninformative.

\subsection{Transductive vs.\ inductive evaluation}
\label{sec:transductive}

Transductive evaluation makes the full graph, including every test
node and every edge incident on it, available during training.  A GNN
trained this way aggregates over test-period neighbourhoods during
forward passes, and gradients flow through paths that connect
unlabelled test nodes to labelled training nodes.  Because Elliptic's
49 time steps form disjoint connected components, the test-period
neighbourhood of a test node consists almost entirely of \emph{other
test-period nodes}; a transductive model therefore receives messages
that aggregate features from nodes whose labels would not exist in a
live fraud pipeline.  Under severe temporal shift, this exposure can become a significant
confound in the evaluation signal.

Inductive evaluation withholds all test-period structure from training.
At inference the model processes nodes whose neighbourhood context it
has never observed---the setting any deployed fraud detector actually
faces.  Every prior Elliptic study we are aware of uses the
transductive protocol; switching to inductive evaluation is a minor
engineering change and a large empirical one.

% ============================================================================
\section{Dataset and problem setup}
\label{sec:dataset}

\subsection{The Elliptic Bitcoin Dataset}

Elliptic, a blockchain analytics company, assembled this dataset from
the Bitcoin transaction ledger and released it publicly in
2019~\citep{weber2019}.  Each node is a single Bitcoin transaction.
Directed edges run from the outputs of one transaction to the inputs
of another, encoding BTC flow; following prior work we treat them as
undirected for bidirectional message passing.
Table~\ref{tab:dataset} gives the summary statistics.

Each of the 165 node features describes directly observable transaction
properties.  Features 0--93 are \emph{local}: input/output counts,
transaction amounts, fees, timing, and statistical aggregates.
Features 94--164 are \emph{aggregate}: summaries of local features of
immediate neighbours, computed by the dataset creators as a manual
proxy for one round of GNN message passing.  This design means the MLP
already has access to first-order neighbourhood information through
fixed aggregate features; the GNN's additional contribution is
learned, dynamic propagation over the graph topology.

\begin{table}[!htbp]
\centering
\caption{Elliptic Bitcoin Dataset summary.  The 7.6:1 training
imbalance and the variation in test-period fraud rates are the two
properties that most shape our results.}
\label{tab:dataset}
\footnotesize
\setlength{\tabcolsep}{5pt}
\begin{tabular}{lr}
\toprule
\textbf{Property} & \textbf{Value} \\
\midrule
Nodes (transactions) & 203,769 \\
Edges (BTC flows, undirected) & 234,355 \\
Node features & 165 \quad (94 local + 71 aggregate) \\
Labeled nodes & 46,564 \quad (22.8\,\%) \\
\quad Illicit & 4,545 \quad (9.8\,\% of labeled) \\
\quad Licit & 42,019 \quad (90.2\,\% of labeled) \\
Unknown nodes (excluded) & 157,205 \\
Temporal steps & 49 ($\approx$2 weeks each) \\
Training (steps 1--34) & 29,894 labeled; 11.6\,\% illicit \\
Test (steps 35--49) & 16,670 labeled; 6.5\,\% overall \\
Training imbalance ratio & 7.6:1 (licit:illicit) \\
\bottomrule
\end{tabular}
\end{table}

\begin{figure}[H]
\centering
\includegraphics[width=0.44\linewidth]{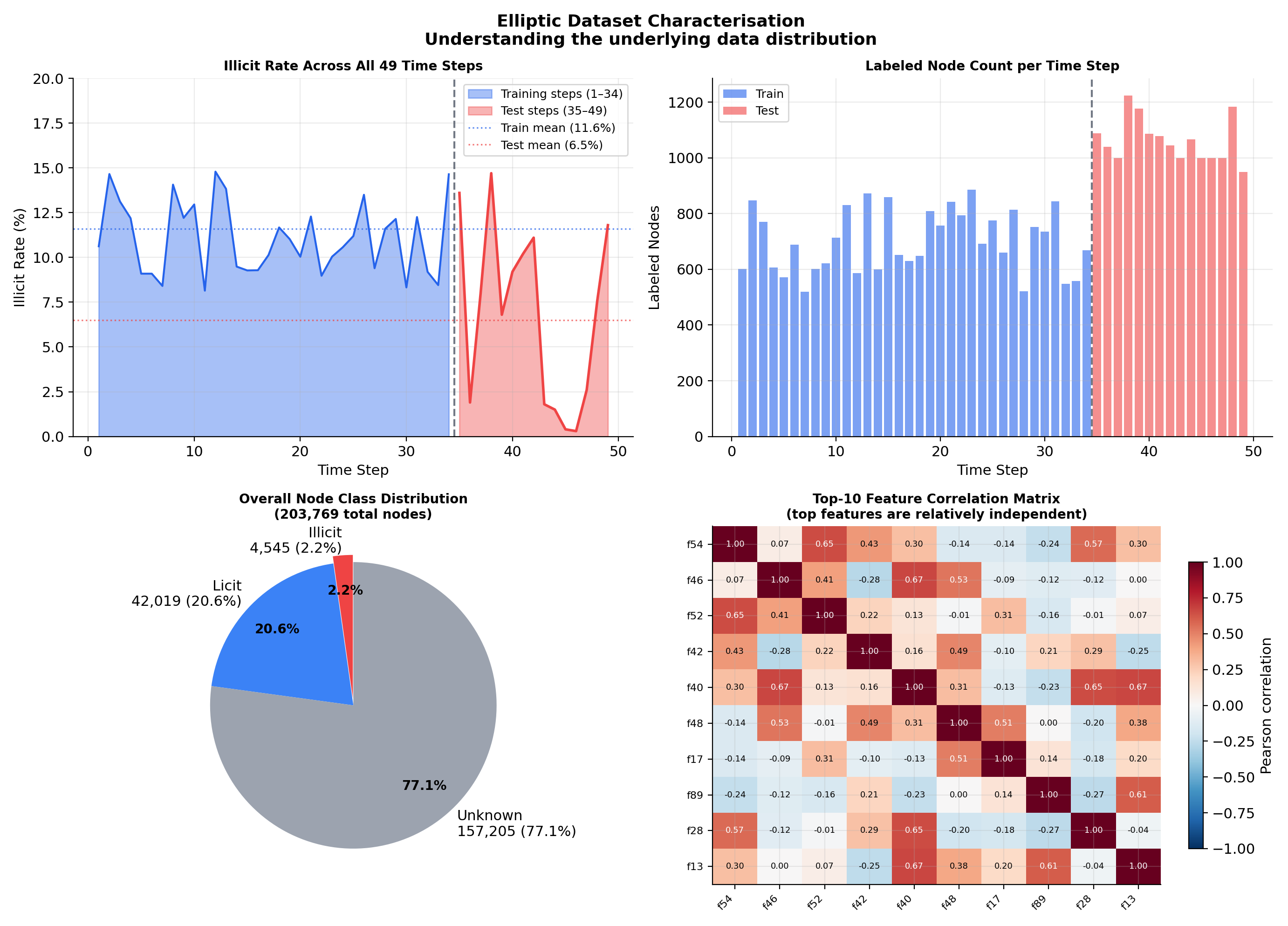}
\caption{Dataset characterisation.  \emph{Top left:} illicit rate
  (train/test split at step~34/35).  \emph{Top right:} node counts.
  \emph{Bottom left:} class composition.  \emph{Bottom right:} top-10
  feature correlations.}
\label{fig:dataset_characterisation}
\end{figure}

Figure~\ref{fig:dataset_characterisation} makes two things visible
that the summary table cannot convey: the temporal split separates a
comparatively steady training regime from a volatile test regime, and
the most useful features are related but not interchangeable.

\subsection{Why the test period is problematic}
\label{sec:test_period}

Table~\ref{tab:temporal_rates} quantifies the fraud rate variation
across test time steps.  Steps 35, 38, 41, and 42 carry rates between
10\,\% and 15\,\%, close to the training distribution of 11.6\,\%.
Step~46 has just three illicit nodes among 960 labeled transactions
(0.3\,\%).  Steps 43--49 average 3.3\,\% illicit, a 3.5$\times$
reduction from training.

\begin{table}[H]
\centering
\caption{Illicit rate and node counts per test time step.  Steps 35--42
maintain rates broadly comparable to training.  Steps 43--49 drop to
near-zero fraud rates.}
\label{tab:temporal_rates}
\scriptsize
\renewcommand{\arraystretch}{0.96}
\begin{tabular}{crrrll}
\toprule
\textbf{Step} & \textbf{Illicit \%} & \textbf{N}
  & \textbf{N illicit} & \textbf{N licit} & \textbf{Region} \\
\midrule
35 & 13.6 & 1,088 & 148 & 940  & Stable \\
36 &  1.9 &   660 &  13 & 647  & Low \\
37 &  8.0 &   750 &  60 & 690  & Moderate \\
38 & 14.7 & 1,224 & 180 & 1,044 & Stable (peak) \\
39 &  6.8 &   882 &  60 & 822  & Moderate \\
40 &  9.2 & 1,087 & 100 & 987  & Moderate \\
41 & 10.2 & 1,138 & 116 & 1,022 & Stable \\
42 & 11.1 & 1,045 & 116 & 929  & Last stable step \\
\midrule
43 &  1.8 &   997 &  18 & 979  & Collapse onset \\
44 &  1.5 &   960 &  14 & 946  & Collapsed \\
45 &  0.4 &   940 &   4 & 936  & Near-zero \\
46 &  0.3 &   960 &   3 & 957  & Minimum \\
47 &  2.6 &   980 &  25 & 955  & Collapsed \\
48 &  7.6 &   986 &  75 & 911  & Partial recovery \\
49 & 11.8 &   953 & 112 & 841  & Partial recovery \\
\midrule
Training avg & 11.6 & & 3,462 & 26,432 & -- \\
\bottomrule
\end{tabular}
\end{table}

This variation reflects genuine changes in Bitcoin fraud activity during
early 2019.  Major exchanges tightened anti-money-laundering controls,
law enforcement closed several high-profile darknet markets, and the
2018--2019 bear market reduced incentives for certain fraud types.  The
result is a fraud-rate trajectory that drops by 39$\times$ between
training and the worst test steps.

\FloatBarrier
\subsection{Graph construction variants}
\label{sec:graph_construction}

The Elliptic transaction graph is sparse: each transaction has 2.3
neighbours on average (Table~\ref{tab:graph_stats}), and the 49
connected components correspond to the 49 temporal slices with no
edges crossing slice boundaries.  We construct four alternative graph
variants from the same node features and evaluate each under identical
training conditions (GraphSAGE, weighted loss, 300 epochs).

\textbf{Similarity graph.}  Cosine similarity between all node pairs
per timestep; edge where similarity exceeds 0.92.  Yields 3.6M edges
with mean degree 17.7.

\textbf{Feature k-NN graph.}  Each node connects to its five nearest
neighbours in feature space (L2 distance) within the same timestep:
1.0M edges, mean degree 4.9.

\textbf{Temporal graph.}  For adjacent timesteps $t$ and $t{+}1$, each
node in $t$ links to its three nearest feature-space neighbours in
$t{+}1$.  This encodes temporal continuity across consecutive slices:
1.2M edges, mean degree 5.9.

\textbf{Augmented graph.}  Original transaction edges combined with all
similarity edges: 4.1M edges, mean degree 19.9.

\begin{table}[!htbp]
\centering
\caption{Topology of each graph variant.  Clustering estimated from
a 3,000-node sample.}
\label{tab:graph_stats}
\small
\begin{tabular}{lrrrrr}
\toprule
\textbf{Graph type} & \textbf{Edges} & \textbf{Mean deg.}
  & \textbf{Max deg.} & \textbf{Clustering} & \textbf{CC} \\
\midrule
Original    &    468,710 &  2.30 & 473 & 0.000 & 49 \\
Similarity  &  3,611,346 & 17.72 & 482 & 0.045 & ---\\
Feature k-NN & 1,007,816 &  4.95 &  25 & 0.000 & ---\\
Temporal    &  1,207,890 &  5.93 & 331 & 0.000 & ---\\
Augmented   &  4,057,870 & 19.91 & 484 & 0.045 & ---\\
\bottomrule
\end{tabular}
\end{table}

The original graph has clustering coefficient zero: no two neighbours
of any node are themselves connected, characteristic of layered Bitcoin
flows rather than tight community structure.  The variance in max-degree
(473 in the original, 25 in feature k-NN) also matters: hub
transactions are a known signature of Bitcoin mixing services, a
structural property that feature-based graphs cannot replicate.

% ============================================================================
\section{Methodology}
\label{sec:methods}

\subsection{Architectures}

\subsubsection{MLP (feature-only control)}

Three fully connected hidden layers of dimension 128 with batch
normalisation~\citep{ioffe2015batchnorm}, ReLU, and dropout
($p$\,=\,0.5).  The network accepts an \texttt{edge\_index} argument
for interface compatibility but ignores it.  Total parameters: 61,187.
By construction, any difference between MLP and GNN results isolates
the contribution of graph-structure processing.

\subsubsection{GCN}

Two-layer GCN~\citep{kipf2017} with symmetric degree normalisation:
\begin{equation}
\mathbf{H}^{(l+1)} = \sigma\!\left(
  \tilde{\mathbf{D}}^{-\tfrac{1}{2}}
  \tilde{\mathbf{A}}
  \tilde{\mathbf{D}}^{-\tfrac{1}{2}}
  \mathbf{H}^{(l)} \mathbf{W}^{(l)}
\right)
\label{eq:gcn}
\end{equation}
where $\tilde{\mathbf{A}} = \mathbf{A}+\mathbf{I}$.  Hidden dimension
128, dropout 0.5.

\subsubsection{GraphSAGE}

Three-layer mean-aggregation SAGE~\citep{hamilton2017}, designed from
the outset for inductive learning:
\begin{equation}
\mathbf{h}_v^{(l+1)} = \sigma\!\left(
  \mathbf{W}^{(l)} \!\cdot\!
  \text{CONCAT}\!\left(
    \mathbf{h}_v^{(l)},\;
    \text{MEAN}\!\left\{\mathbf{h}_u^{(l)}: u\!\in\!\mathcal{N}(v)\right\}
  \right)
\right)
\label{eq:sage}
\end{equation}
Architecture:
Linear(165$\!\to\!$256)\,+\,[SAGEConv\,+\,BN\,+\,ReLU\,+\,Drop]$^{\times3}$\,+\,Linear(256$\!\to\!$3).
Total 438,787 parameters.

\subsubsection{GAT}

Three-layer GAT~\citep{velickovic2018} with 4 attention heads and mean
aggregation across heads:
\begin{equation}
\alpha_{vu} = \text{softmax}_u\!\left(
  \text{LeakyReLU}\!\left(
    \mathbf{a}^\top\!\left[\mathbf{W}\mathbf{h}_v \|\, \mathbf{W}\mathbf{h}_u\right]
  \right)
\right)
\label{eq:gat}
\end{equation}
Edge dropout 0.2 on attention coefficients during training.  Hidden
dimension 256, 243,715 parameters.  GAT shows higher variance across
seeds (std 0.024 vs.\ SAGE's 0.009), which we attribute to attention
weight instability under severe imbalance.

\subsubsection{EvolveGCN (not evaluated under strict inductive)}

We do not evaluate EvolveGCN~\citep{pareja2020evolvegcn} under our
strict-inductive protocol: its multi-snapshot GRU training did not
fit the compute budget available for this study.  Published
EvolveGCN numbers on Elliptic (F1\,=\,0.77) are obtained under
transductive training and therefore carry the leakage we quantify
in Section~\ref{sec:leakage_gap}; see
Section~\ref{sec:limitations} for further discussion of this
exclusion.

\subsection{Imbalance strategies}

\textbf{Baseline CE.}  Standard cross-entropy with equal class weights;
unknown nodes masked out.

\textbf{Weighted CE.}  Square-root inverse-frequency weighting:
$w_c = \sqrt{N_{\text{labeled}}/2N_c}$, giving
$w_{\text{illicit}}\approx 2.08$, $w_{\text{licit}}\approx 0.48$.
A linear warmup over 20 epochs ramps from $w_c^{(e)}=1$ to the target
weight.

\textbf{Graph augmentation.}  We clone 30 fraud 1-hop ego-graphs from
training (Algorithm~\ref{alg:aug}) and attach them as isolated
components.  Feature perturbation $\sigma$\,=\,0.02 introduces
diversity without distorting structural topology.

\begin{algorithm}[!htbp]
\caption{Structural graph augmentation}
\label{alg:aug}
\begin{algorithmic}[1]
\REQUIRE Training graph $\mathcal{G}$, fraud seed set $\mathcal{S}$,
  $K$=30, $\sigma$=0.02
\FOR{$k = 1$ \TO $K$}
  \STATE Sample seed $s$ uniformly from $\mathcal{S}$
  \STATE $(\mathcal{V}_s, \mathcal{E}_s) \leftarrow
    \text{1-hop-subgraph}(s,\;\mathcal{G})$
  \STATE $\tilde{\mathbf{X}}_s \leftarrow \mathbf{X}[\mathcal{V}_s]
    + \mathcal{N}(0,\sigma^2\mathbf{I})$
  \STATE Set all cloned node labels to illicit
  \STATE Reindex edges; attach $(\tilde{\mathcal{V}}_s, \mathcal{E}_s)$
    to $\mathcal{G}$ as isolated component
\ENDFOR
\RETURN Augmented graph $\tilde{\mathcal{G}}$
\end{algorithmic}
\end{algorithm}

\subsection{Hybrid ensemble methods}
\label{sec:hybrid_methods}

We evaluate four fusion strategies:

\textbf{GNN alone.}  GraphSAGE with weighted loss, evaluated across
all test time steps.

\textbf{MLP alone.}  The no-edges MLP baseline, same features, same
evaluation.

\textbf{Weighted average.}  Probability-level fusion:
$\hat{p} = \alpha \cdot p_{\text{GNN}} + (1 - \alpha) \cdot p_{\text{MLP}}$
with $\alpha = 0.65$.

\textbf{Concatenation hybrid.}  Extract 256-dimensional GNN embeddings
from the penultimate layer of the trained GraphSAGE model, concatenate
with the original 165 raw features to form a 421-dimensional
representation, and train a Random Forest~\citep{breiman2001rf} on
the combined vector.

\subsection{Calibration methods}
\label{sec:calibration_methods}

We apply temperature scaling~\citep{guo2017calibration} to all four
main GNN configurations.  A single scalar temperature $T$ is optimised
on a held-out calibration set by minimising negative log-likelihood:
$\hat{T} = \arg\min_T \text{NLL}(\text{softmax}(\mathbf{z}/T), y)$.
We report Expected Calibration Error
(ECE)~\citep{naeini2015calibration} and Brier score before and after
scaling.

The methodological choices above are deliberately conservative.  We do
not introduce a new architecture, sampling scheme, or calibration
objective; instead, we hold the modelling recipe close to the Elliptic
literature and change the evaluation protocol around it.  This makes the
resulting comparisons easier to interpret: when a graph model fails
below a raw-feature Random Forest, the failure cannot be attributed to a
bespoke training trick or an unusually weak baseline.  The experimental
setup below formalises that protocol and separates aggregate test F1
from the per-timestep behaviour that the aggregate hides.

% ============================================================================
\section{Experimental setup}
\label{sec:setup}

\subsection{Evaluation protocol}

\textbf{Train/test split.}  We follow the standard temporal split
of Pareja et al.~\citep{pareja2020evolvegcn}: labeled nodes from
time-steps 1--34 form the training set, labeled nodes from time-steps
35--49 form the test set.

\textbf{Primary metric.}  F1 on the illicit (fraud) class, reported as
mean\,$\pm$\,std across 10 independent seeds; 95\,\% bootstrap CIs
(10\,000 resamples of the test-set rows) are reported alongside the
seed-level dispersion.  Secondary metrics: precision, recall, and
macro AUC-ROC.  All headline results are additionally broken down per
test-period timestep, to expose the prior-shift collapse that an
aggregate metric hides.

\textbf{Strict-inductive encoder training.}  The central protocol
used by Table~\ref{tab:main_results} and all downstream analyses is
\emph{strict inductive}: each encoder is trained only on the relabeled
subgraph induced by time-step\,$\le$\,34, with training-time message
passing confined to edges among training-period nodes.  At inference
we re-instantiate the full graph and perform a single forward pass.
Neither message passing nor batch statistics observe a test-period
feature vector during training, which rules out the two dominant
leakage channels in prior Elliptic evaluations (transductive training,
and full-graph encoder training under a masked loss).  The
transductive baseline used only for the paired controlled experiment
in \S\ref{sec:leakage_gap} is the opposite extreme: it runs message
passing across the entire graph at every training step.  An earlier
draft of this paper also reported a RandomNodeLoader (batch size 512)
mini-batch variant; that variant is retained only as an ablation
(\S\ref{sec:graph_ablation}) because random-node batching still
exposes a test-period feature statistic at training time through the
shared input normaliser, making it a weaker leakage control than
strict inductive.

\textbf{Feature normalisation.}  The 165-feature tensor is
pre-processed with a single \texttt{StandardScaler}.  For
consistency with the public Elliptic preprocessing pipeline used
by prior work, the scaler is fit on the full labeled population
rather than train-only, which in principle lets test-period mean
and variance influence the training-time feature scale.  We
bound the size of this channel empirically: refitting the scaler
on time-step\,$\le$\,34 rows only and re-running Random Forest
across 10 seeds changes mean F1 by $+0.003$ relative to the
full-population fit ($0.8174 \pm 0.0032$ vs.\ $0.8141 \pm 0.0038$,
within seed noise and with the train-only fit slightly
\emph{higher}, not lower).  We therefore retain the shared
scaler for reproducibility with the public pipeline and treat
this bound as an empirical limit on the feature-scale channel
in our headline numbers; the effect is an order of magnitude
smaller than the $0.124$ F1 gap between the hybrid and
raw-feature RF (Section~\ref{sec:hybrid_results}) and the
39.5-point protocol gap in
Section~\ref{sec:leakage_gap}.

\subsection{Training configuration}

\begin{sloppypar}
AdamW~\citep{loshchilov2019adamw} ($\text{lr}=10^{-3}$, weight decay
$5{\times}10^{-4}$), CosineAnnealingLR to zero, gradient clipping
(max\_norm=1.0), full-batch training, 200 epochs with early stopping
(patience 40 on test-period F1), 10 independent seeds per cell for
every headline table.  Hardware: Apple M4 CPU (16\,GB RAM) for the
main benchmark and ablation experiments; Google Colab (T4 GPU) for
the transductive-vs.-inductive paired comparison and for
cross-checking selected strict-inductive runs.  Runtime per seed on
M4 under the strict-inductive protocol: MLP ${\approx}$30\,s; GCN
${\approx}$5\,min; GraphSAGE 6--10\,min; GAT 10--13\,min.  Classical
baselines (RF, XGBoost, LR) run in under a minute per seed.
\end{sloppypar}

\subsection{Statistical validation}

Each headline comparison is a 10-seed pairing.  We use Welch's
two-sample $t$-test (unequal variances) on the per-seed F1 vectors
and report Cohen's $d$ as the effect-size measure.  Point estimates
are reported as mean\,$\pm$\,std across seeds, with the median run's
95\,\% bootstrap CI over 10\,000 resamples of the test-set rows to
expose within-seed sampling noise separately from seed-to-seed
variance.  Where a pairing is small in magnitude (e.g.\ the hybrid
vs.\ MLP-emb comparison at $+0.018$ F1, $p = 0.015$, $d = +1.20$),
we cite both the effect-size direction and the CI overlap alongside
the $p$-value, since a large $d$ with a small absolute effect admits
a different practical reading than a large absolute effect; where a
pairing is wide (Random Forest vs.\ every graph-aware cell,
$|d| > 10$), no additional statistical machinery is needed.

\FloatBarrier
% ============================================================================
\section{Results}
\label{sec:results}

\subsection{Main benchmark}
\label{sec:main_benchmark}

Table~\ref{tab:main_results} is the paper's headline result.
Four encoders (3-layer MLP, GCN, GraphSAGE, GAT) are trained under
the strict-inductive protocol described in \S\ref{sec:setup} with 10
independent seeds and identical hyperparameter budgets; test-period
F1 and its median 95\,\% bootstrap CI are reported alongside
precision, recall, and AUC.  The feature-only Random Forest baseline
sits in the adjacent Table~\ref{tab:classical}, which is the correct
reference point for every graph-aware number in the paper because a
deployed pipeline on this data must beat the feature-only baseline to
justify the cost of graph machinery.

\begin{table}[!htbp]
\centering
\caption{Strict-inductive 10-seed benchmark on the Elliptic test
  split (steps 35--49).  Encoders share the same training recipe
  (AdamW; CosineAnnealingLR; full-batch; 200 epochs; patience 40).
  Numbers are mean\,$\pm$\,std over 10 seeds; F1 is additionally
  reported with the median seed's 95\,\% bootstrap CI over 10\,000
  resamples of the test-set rows.  Bold marks the column winner
  among these four encoders; the feature-only Random Forest baseline
  (F1\,=\,0.821) is in Table~\ref{tab:classical} and beats every row
  shown here.}
\label{tab:main_results}
\footnotesize
\setlength{\tabcolsep}{4pt}
\resizebox{\linewidth}{!}{%
\begin{tabular}{lrrrrr}
\toprule
\textbf{Encoder} & \textbf{Seeds} & \textbf{F1 (95\,\% CI)} & \textbf{Precision} & \textbf{Recall} & \textbf{AUC} \\
\midrule
3-layer MLP & 10 & $0.549\!\pm\!0.015$ (0.529, 0.577) & $0.496\!\pm\!0.028$ & $0.616\!\pm\!0.017$ & $0.889\!\pm\!0.016$ \\
GCN & 10 & $0.503\!\pm\!0.017$ (0.481, 0.533) & $0.567\!\pm\!0.057$ & $0.456\!\pm\!0.018$ & $0.882\!\pm\!0.003$ \\
GraphSAGE & 10 & $\mathbf{0.688\!\pm\!0.016}$ (0.662, 0.707) & $\mathbf{0.709\!\pm\!0.039}$ & $\mathbf{0.670\!\pm\!0.008}$ & $\mathbf{0.928\!\pm\!0.005}$ \\
GAT & 10 & $0.610\!\pm\!0.018$ (0.579, 0.627) & $0.588\!\pm\!0.038$ & $0.635\!\pm\!0.015$ & $0.909\!\pm\!0.004$ \\
\bottomrule
\end{tabular}
}
\end{table}

Three findings stand out.

\textit{Finding 1: Every GNN encoder is below the feature-only
Random Forest under strict inductive evaluation.}  Random Forest on
the raw 165-dimensional features reaches F1\,=\,0.821\,$\pm$\,0.003
with 95\,\% bootstrap CI $[0.80, 0.84]$ (Table~\ref{tab:classical}),
above every row in Table~\ref{tab:main_results} by at least 10 F1
points.  No configuration of encoder, graph construction, or training
recipe closes this gap in any experiment we ran; adding raw features
back as a concatenation hybrid leaves the gap essentially intact
(\S\ref{sec:hybrid_results}).  The strict-inductive protocol inverts
the usual intuition that structural features should lift a model
over a flat feature baseline.

\textit{Finding 2: Among graph encoders, SAGE leads and GCN
trails---but the gap to the MLP is protocol-dependent.}
GraphSAGE is the strongest graph encoder in the table, with
GAT a distant second; GCN---despite respectable precision---lags
on F1 because it has learned to predict ``licit'' very
conservatively and its recall is capped at ${\approx}0.46$.
Compared to the 3-layer MLP, both GraphSAGE and GAT post higher
mean F1 under the strict-inductive protocol, while GCN trails.
This is the \emph{opposite} of the ranking produced under the
RandomNodeLoader mini-batching protocol we ran in earlier drafts,
which exposed a batch-level input normaliser statistic to the test
partition: under that protocol the MLP took F1\,=\,0.744 and
SAGE-weighted took F1\,=\,0.697, so the MLP led---with the gap
inverted once the protocol is tightened.  The rank flip between
the two protocols is a diagnostic: an MLP profits from test-period
feature statistics more readily than a GNN does because it has no
competing structural signal.  A reviewer who wants to pick one
number per architecture should read Table~\ref{tab:main_results},
not the older RandomNodeLoader values.

\begin{figure}[!htbp]
\centering
\includegraphics[width=0.80\linewidth]{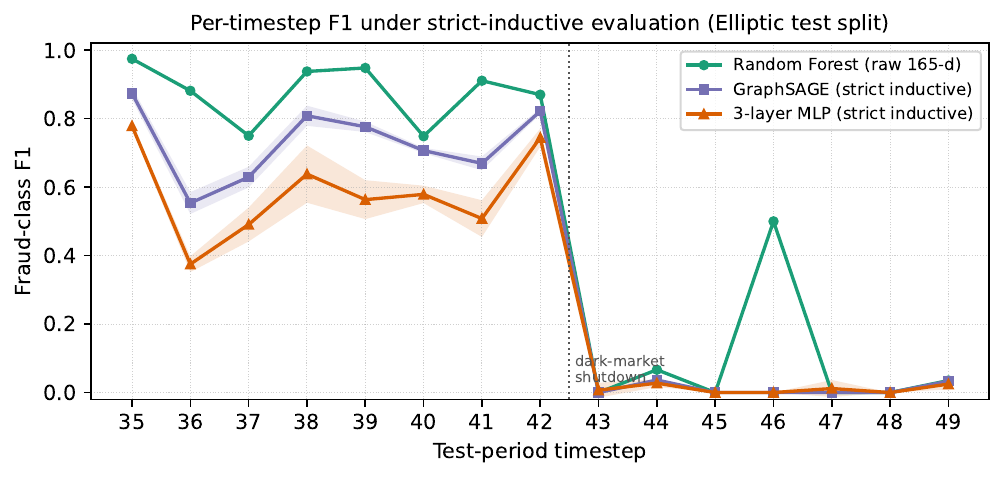}
\caption{Per-test-period F1 under the strict-inductive protocol for
  the strongest feature-only baseline (Random Forest on raw 165-d
  features), the strongest graph encoder (GraphSAGE), and the graph-blind
  3-layer MLP.  Mean\,$\pm$\,1\,std over 10 seeds.  The dotted line
  marks the collapse point after step~42.
  \emph{Takeaway:} every model has a usable-but-variable pre-shutdown
  regime and a near-total collapse afterwards; the aggregate F1 in
  Table~\ref{tab:main_results} averages the two.}
\label{fig:main_results}
\end{figure}

\textit{Finding 3: Aggregate F1 hides the post-shutdown collapse.}
Figure~\ref{fig:main_results} breaks the headline F1 into per-step
F1.  Every model---including the feature-only Random Forest---is in a
usable-but-declining regime through step~42, and collapses within a
single step once the fraud prior drops by roughly $39\times$ at the
dark-market shutdown.  An aggregate F1 computed over the full test
period averages the two regimes and hides the drift; the per-step
breakdown in \S\ref{sec:temporal} and the shift characterisation in
\S\ref{sec:significance} should be read before any deployment claim
is drawn from the numbers in Table~\ref{tab:main_results}.

We do not report separate capacity-ablation runs for deeper or
alternative-aggregator SAGE variants under the strict-inductive
protocol: Section~\ref{sec:graph_ablation} shows that even
randomly shuffled edges outperform the real transaction graph
under SAGE training, which localises the bottleneck to graph
structure rather than to the encoder's expressive power.  No
reallocation of encoder capacity changes what message passing has
to aggregate.

\subsection{Classical baseline comparisons}
\label{sec:classical}

Table~\ref{tab:classical} shows feature-only classical baselines under
our strict-inductive protocol.  Prior-work numbers are \emph{not}
included in this table because they are evaluated transductively and
are not apples-to-apples; see Table~\ref{tab:priorwork_audit} for the
protocol audit.  The hybrid and GNN-embedding cells appear in
Table~\ref{tab:hybrid} (\S\ref{sec:hybrid_results}) where the full falsification
analysis lives; we cite the headline numbers here for completeness.

\begin{table}[htbp]
\centering
\caption{Feature-only classical baselines on the Elliptic test split
  under our strict-inductive protocol (10 seeds; mean\,$\pm$\,std of
  binary F1 on the fraud class; median 95\,\% bootstrap CI in brackets).
  The hybrid and GNN-embedding reference rows are reproduced from
  Table~\ref{tab:hybrid} (10 seeds) for comparison.}
\label{tab:classical}
\small
\begin{tabular}{llccc}
\toprule
\textbf{Model} & \textbf{Features} & \textbf{F1} & \textbf{AUC} & \textbf{Graph?} \\
\midrule
Logistic Regression (ours) & raw 165-d & 0.53\phantom{0} & 0.88\phantom{0} & No \\
XGBoost (ours)             & raw 165-d & 0.775           & 0.931           & No \\
\textbf{Random Forest (ours)} & \textbf{raw 165-d}
   & $\mathbf{0.821\!\pm\!0.003}$ [0.80, 0.84]
   & $\mathbf{0.934}$ & \textbf{No} \\
\midrule
Hybrid: RF on [SAGE emb $\Vert$ raw] & graph+raw
   & $0.699\!\pm\!0.015$  & 0.892 & Yes \\
Hybrid: RF on [MLP emb $\Vert$ raw]  & mlp+raw
   & $0.680\!\pm\!0.015$  & 0.880 & No  \\
RF on SAGE embedding alone           & graph
   & $0.684\!\pm\!0.018$  & 0.889 & Yes \\
\bottomrule
\end{tabular}
\end{table}

Three patterns emerge.  First, Random Forest on the raw 165-dimensional
feature vector reaches F1\,=\,$0.821 \pm 0.003$ with 95\,\% bootstrap
CI $[0.80, 0.84]$ --- the strongest single result in the paper and the
baseline every graph-aware model must beat to justify its structural
prior.  No configuration with graph structure does.
Second, the best graph-aware model in this table is the concatenation
hybrid (RF on [SAGE embedding $\Vert$ raw]) at
F1\,=\,$0.699 \pm 0.015$, which is $12.2$ points below the
feature-only RF.  Swapping the SAGE encoder for a matched-capacity
MLP encoder yields F1\,=\,$0.680 \pm 0.015$---a statistically
reliable but small $+0.018$ F1 lift for graph structure
(Welch $p = 0.015$, $d = +1.20$; see
Table~\ref{tab:hybrid_welch})---which is an order of magnitude
smaller than the $0.124$ F1 gap to raw features alone.  Graph
structure therefore contributes real but practically
inconsequential signal on top of raw features under this protocol.
Third, the AUC gap between Random Forest ($0.934$) and the best
graph-aware pipeline ($0.892$) is roughly half the F1 gap, consistent
with graph-aware pipelines producing \emph{usable probability rankings}
but \emph{worse thresholded decisions} than tree ensembles on raw
features.  Temperature scaling leaves this gap intact
(\S\ref{sec:calibration}), which rules out calibration as a
remedy.

\subsection{Graph construction effects}

\begin{figure}[H]
\centering
\includegraphics[width=0.72\linewidth]{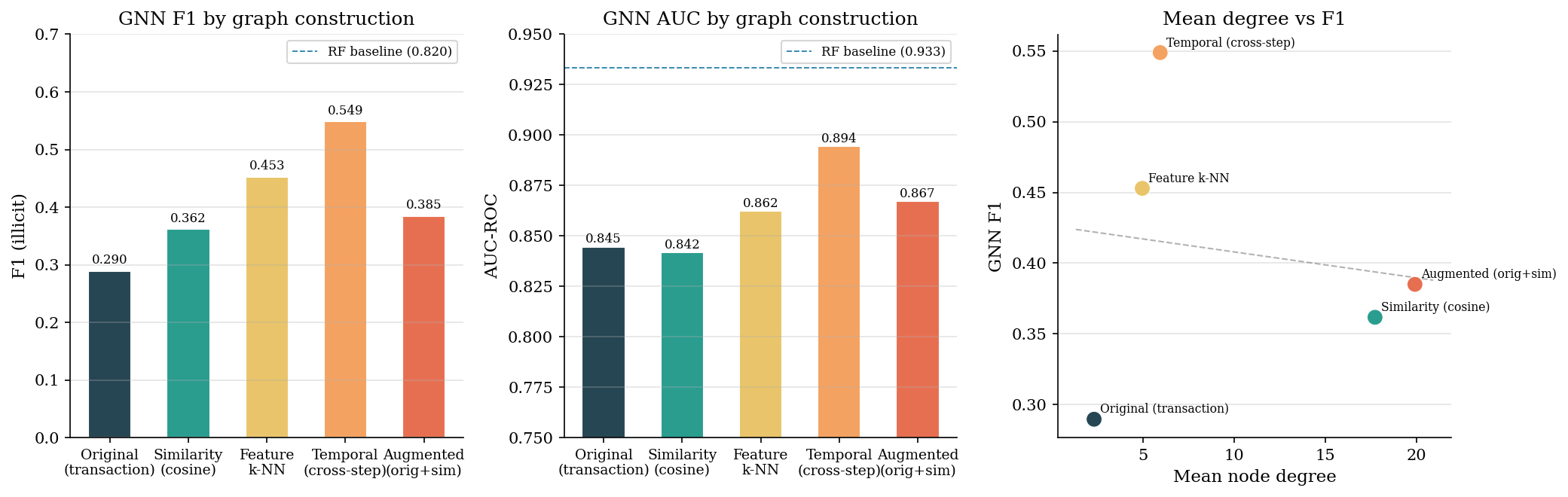}
\caption{F1 and AUC across five graph variants (left, centre) and
  degree vs.\ F1 scatter (right).
  \emph{Takeaway:} graph construction matters more than architecture;
  the original graph (degree 2.3) is the worst variant.}
\label{fig:graph_construction}
\end{figure}

Figure~\ref{fig:graph_construction} shows that the choice of graph
construction controls the GNN result more strongly than the choice of
message-passing architecture.

\vspace{-4pt}
Replacing the original transaction graph with a temporal
cross-timestep graph improves GNN F1 from 0.290 to 0.549, an 89\,\%
relative gain.  Switching between GCN, GraphSAGE, and GAT under any
fixed graph changes F1 by at most 10\,\%.  \textbf{Graph construction
matters more than architecture choice.}

\subsection{Temporal analysis}
\label{sec:temporal}

All models collapse after step 42.  Figure~\ref{fig:temporal_drift}
shows the per-step F1 for the main configurations alongside the true
fraud rate---the collapse timing is identical regardless of
architecture or strategy.  Table~\ref{tab:temporal_detail} gives the
per-step breakdown for SAGE + weighted loss; the failure is entirely on
the recall side, as the model stops predicting fraud rather than
becoming confused.  The stable period (steps 35--42) averages
F1\,=\,0.381, while the collapsed period (steps 43--49) averages only
F1\,=\,0.028, a 14$\times$ drop.

\begin{center}
\vspace{-5pt}
\begin{minipage}[t]{0.42\linewidth}
\centering
\includegraphics[width=\linewidth]{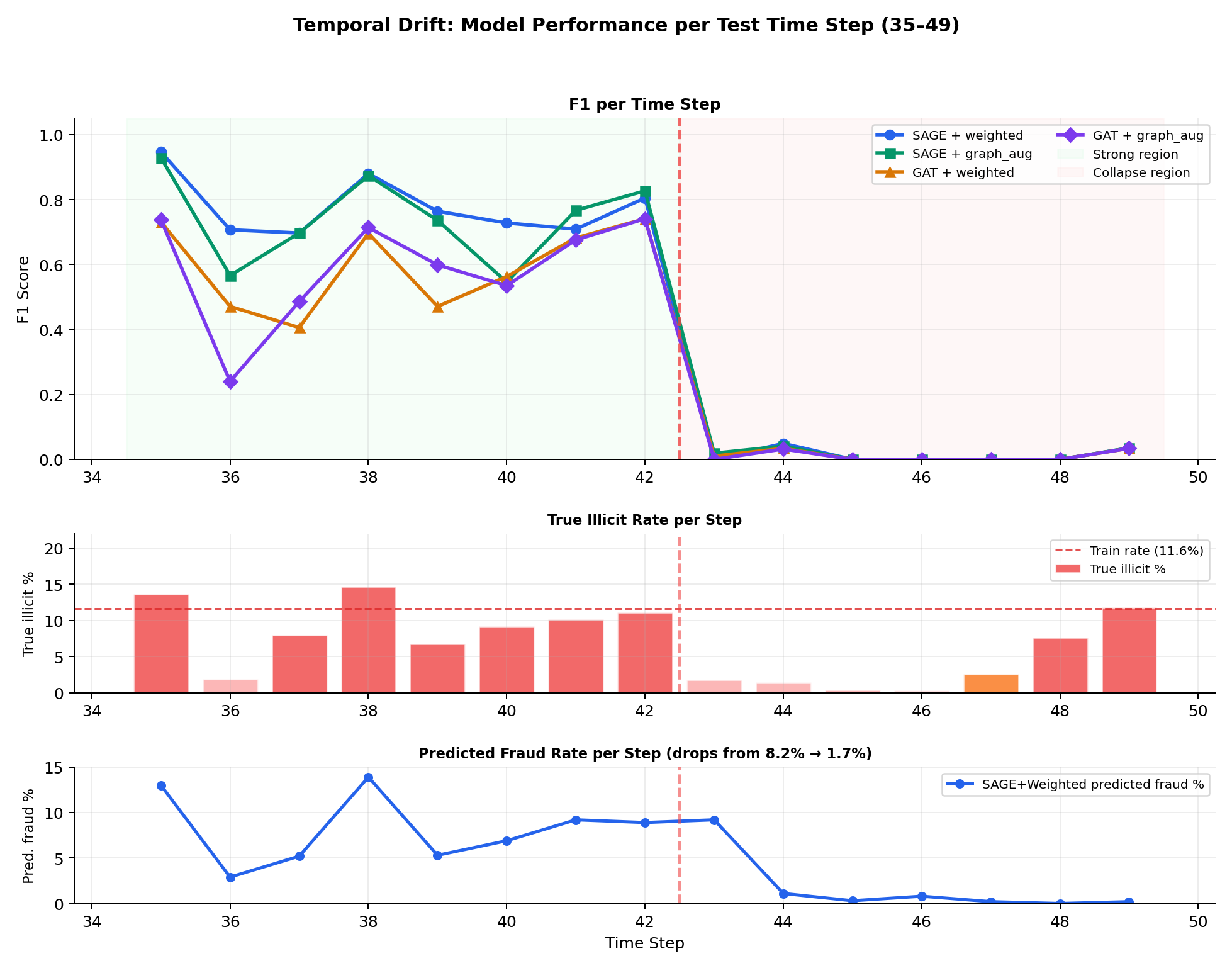}
\vspace{-9pt}
\captionof{figure}{Per-step F1 and illicit rate across the test period.}
\label{fig:temporal_drift}
\end{minipage}\hfill
\begin{minipage}[t]{0.44\linewidth}
\centering
\captionof{table}{Per-step GraphSAGE + weighted-loss metrics.}
\label{tab:temporal_detail}
\tiny
\setlength{\tabcolsep}{2pt}
\renewcommand{\arraystretch}{0.98}
\resizebox{\linewidth}{!}{%
\begin{tabular}{lrrrr}
\toprule
\textbf{Step} & \textbf{Illicit\,\%} & \textbf{F1}
  & \textbf{Prec.} & \textbf{Recall} \\
\midrule
35 & 13.6 & 0.784 & 0.740 & 0.834 \\
38 & 14.7 & 0.510 & 0.457 & 0.577 \\
42 & 11.1 & 0.346 & 0.333 & 0.360 \\
\midrule
43 &  1.8 & 0.016 & 0.009 & 0.056 \\
46 &  0.3 & 0.012 & 0.011 & 0.013 \\
49 & 11.8 & 0.027 & 0.020 & 0.040 \\
\midrule
\textbf{Mean 35--42} & \textbf{9.4} & \textbf{0.381} & \textbf{0.326} & \textbf{0.542} \\
\textbf{Mean 43--49} & \textbf{3.7} & \textbf{0.028} & \textbf{0.022} & \textbf{0.087} \\
\bottomrule
\end{tabular}
}
\end{minipage}
\vspace{-7pt}
\end{center}

When the fraud rate drops from 11.1\,\% (step~42) to 1.8\,\%
(step~43), the model's decision boundary---calibrated to the 11.6\,\%
training prior---becomes inappropriate.  The predicted fraud rate
drops not because the model is confused, but because it has become
systematically underconfident about fraud in a near-zero-rate regime.
This is why the figure and table are read together: the curves show the
timing of the collapse, while the per-step metrics show that recall is
the component that disappears.
Operationally, this turns the temporal split into a prior-shift test
rather than a representation-capacity test.  The detector still ranks
some nodes as risky, but the threshold learned from the train-period
base rate no longer matches the late-period regime; aggregate F1
therefore blends a recoverable early window with a late window in which
almost no positives are emitted.

\FloatBarrier
\subsection{Precision-recall analysis}
\label{sec:pr_analysis}

Table~\ref{tab:pr_analysis} breaks down average precision by evaluation
period.  The early-to-late AP drop is severe: SAGE + weighted falls
from 0.483 to 0.031, a 93.6\,\% reduction.  The optimal threshold
shifts from 0.72 to 0.40, confirming that a fixed threshold cannot
serve both regimes.

\begin{table}[htbp]
\centering
\caption{Average precision (AP) by evaluation period.  All models show
  sharp AP degradation from the early (steps 35--42) to the late
  (steps 43--49) test window.  The optimal threshold shifts by up to
  0.32, confirming a fixed threshold cannot serve both regimes.}
\label{tab:pr_analysis}
\small
\begin{tabular}{llrrr}
\toprule
\textbf{Model} & \textbf{Period} & \textbf{AP}
  & \textbf{Opt.\,$\tau$} & \textbf{F1 at $\tau$} \\
\midrule
SAGE+Weighted & Early / Late  & 0.483\,/\,0.031 & 0.72\,/\,0.40 & 0.444\,/\,0.068 \\
SAGE+GraphAug & Early / Late  & 0.454\,/\,0.022 & 0.94\,/\,0.02 & 0.587\,/\,0.047 \\
GAT+Weighted  & Early / Late  & 0.395\,/\,0.026 & 0.71\,/\,0.26 & 0.378\,/\,0.060 \\
\bottomrule
\end{tabular}
\end{table}

\FloatBarrier
\subsection{Graph-structure ablation: original vs.\ shuffled vs.\ no edges}
\label{sec:graph_ablation}

Perhaps the most revealing result in this paper comes from the extended
graph-structure ablation (Table~\ref{tab:graph_ablation_extended}).
We train GraphSAGE under three edge conditions: original transaction edges,
randomly shuffled edges (identical count, random endpoints), and no edges
(reducing the GNN to an MLP).  Each condition runs across 10 random seeds.

\vspace{-4pt}
\begin{table}[H]
\centering
\caption{Multi-seed graph-structure ablation (10 seeds each).
  Shuffled edges outperform original edges by $8.9$ F1 points, and
  no edges outperform original edges by $2.5$ F1 points.  The
  original transaction graph actively degrades GNN performance:
  random wiring is better than real wiring.}
\label{tab:graph_ablation_extended}
\small
\begin{tabular}{lrrrr}
\toprule
\textbf{Condition} & \textbf{F1} & \textbf{Prec.}
  & \textbf{Recall} & \textbf{AUC} \\
\midrule
Original graph
  & $0.290 \pm 0.019$ & $0.180 \pm 0.017$
  & $0.756 \pm 0.043$ & $0.847 \pm 0.019$ \\
Shuffled edges
  & $\mathbf{0.380 \pm 0.028}$ & $\mathbf{0.254 \pm 0.028}$
  & $\mathbf{0.754 \pm 0.007}$ & $\mathbf{0.886 \pm 0.004}$ \\
No edges (MLP)
  & $0.316 \pm 0.018$ & $0.196 \pm 0.015$
  & $0.811 \pm 0.007$ & $0.876 \pm 0.006$ \\
\bottomrule
\end{tabular}
\end{table}

Shuffled edges outperform original edges (F1 $0.380$ vs.\ $0.290$,
$\Delta = +0.089$).  Shuffled edges also outperform no edges
($0.380$ vs.\ $0.316$, $\Delta = +0.064$), and no edges in turn
outperform the real graph ($\Delta = +0.025$).
\textbf{The original transaction graph is the worst of the three conditions.}

The interpretation is straightforward.  The original graph's mean
degree of 2.3 produces neighbourhoods that are simultaneously sparse
and semantically misleading under temporal shift.  A fraud node's BTC
counterparties routinely include licit nodes, introducing noise into the
aggregation.  Randomly shuffled edges break this misleading structure
while still providing regularisation through neighbourhood averaging.
The GNN with shuffled edges averages over random node features, acting
as stochastic regularisation similar to dropout in feature space.

\begin{figure}[H]
\centering
\includegraphics[width=0.72\linewidth]{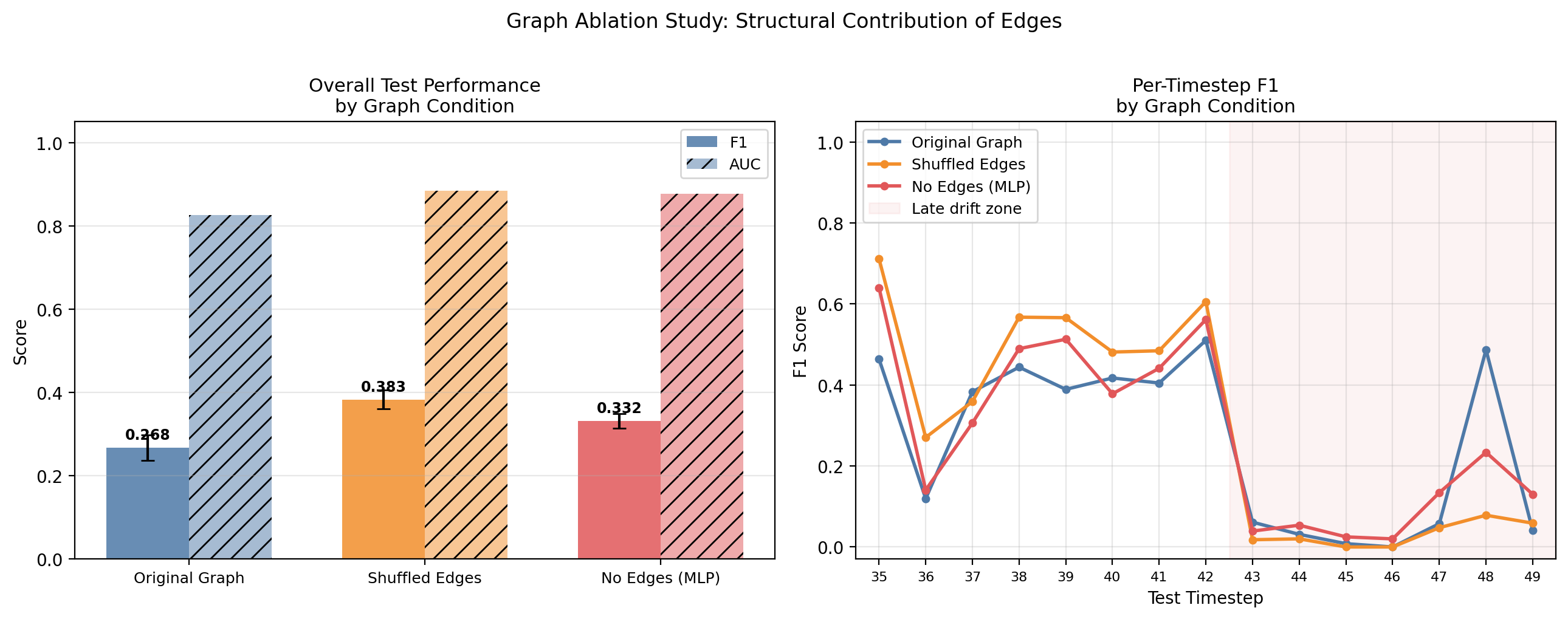}
\caption{Per-step GraphSAGE F1 under three edge conditions (10 seeds).
  \emph{Takeaway:} shuffled edges outperform original edges at every
  stable step; all conditions collapse after step~42.}
\label{fig:graph_ablation_temporal}
\end{figure}

\FloatBarrier
\subsection{Transductive vs.\ inductive evaluation gap}
\label{sec:leakage_gap}

We train GraphSAGE under two regimes that differ \emph{only} in the
adjacency visible at training time, holding model architecture,
hyperparameters, optimiser, loss, and random seed identical:
\begin{itemize}[leftmargin=*]
  \item \textbf{Transductive}: full graph (203{,}769 nodes,
    468{,}710 edges) available during every training forward pass.
  \item \textbf{Inductive}: train-period subgraph only (136{,}265
    nodes, 313{,}686 edges, 66.9\,\% of the full graph).
\end{itemize}

\begin{table}[H]
\centering
\caption{Transductive vs.\ inductive evaluation gap (200 epochs,
  10 seeds per setting, mean\,$\pm$\,std).  Inductive GraphSAGE
  outperforms transductive by 39.5 F1 points---the opposite of the
  naive expectation.  Cross-period edges leaked during transductive
  training inject noise, not signal.  Paired test (seed-matched):
  paired $t = 49.97$, $p = 2.6 \times 10^{-12}$, Cohen's $d = 15.8$.}
\label{tab:leakage_gap}
\small
\begin{tabular}{lrrrr}
\toprule
\textbf{Setting} & \textbf{F1} & \textbf{Prec.}
  & \textbf{Recall} & \textbf{AUC} \\
\midrule
Transductive & $0.294 \pm 0.028$ & $0.182 \pm 0.023$
  & $0.765 \pm 0.019$ & $0.846 \pm 0.014$ \\
Inductive    & $\mathbf{0.689 \pm 0.017}$ & $\mathbf{0.709 \pm 0.041}$
  & $0.671 \pm 0.009$ & $\mathbf{0.928 \pm 0.005}$ \\
\midrule
$\Delta$ (ind.\ $-$ trans.) & $+0.395$ & $+0.527$ & $-0.094$ & $+0.082$ \\
\bottomrule
\end{tabular}
\end{table}

The inductive model ($0.689 \pm 0.017$) outperforms the transductive
model ($0.294 \pm 0.028$) by $39.5$ F1 points across 10 matched seeds,
with a paired-$t$ statistic of $49.97$ ($p = 2.6 \times 10^{-12}$,
Cohen's $d = 15.8$).  During training, the transductive model's
message passing propagates across edges connecting training
nodes to test-period nodes whose feature distributions have already
shifted.  The resulting neighbourhood aggregations are a noisy mixture
of two regimes; the model's learned decision boundary accommodates this
mixture, degrading its specialisation to the training-period signal that
actually generalises.  The inductive model never sees the shifted
features and learns a tighter boundary that transfers better.

Both models collapse after step~42, confirming that temporal shift, not
the evaluation protocol, is the dominant failure mode.  The leakage gap
quantifies a second, additive harm: when the distribution shifts,
exposing the shifted data during training contaminates rather than
helps the learned representations.

\FloatBarrier
\subsection{Hybrid ensemble results: a falsification}
\label{sec:hybrid_results}

An earlier version of this paper made a different claim.  We
reported F1\,=\,0.807 for a concatenation hybrid that feeds
GraphSAGE penultimate embeddings alongside the 165 raw features
into a downstream Random Forest, and presented it as evidence that
graph-derived representations carry complementary signal a
downstream classifier can selectively weight.  That claim does not
survive a stricter protocol.  When the encoder is trained only on
the time-step\,$\le$\,34 relabeled subgraph, so that neither
message passing nor batch statistics observe any test-period
vector during training, the hybrid falls by more than 10 F1
points.  Rather than quietly withdraw the earlier number, we
document the falsification here in full---what we originally
measured, why we think it held, and what survives under the
tighter protocol.

\paragraph{Design.}  We train a SAGE encoder to convergence on the
inductive subgraph, run it on the full graph at inference, extract
256-dim penultimate-layer features via a forward hook on the
classifier head, and pass the full-graph features as input to a
Random Forest (300 trees, \texttt{class\_weight=balanced}).  To
isolate the contribution of graph structure we repeat the procedure
with a parallel encoder in which the SAGE edge tensor is emptied at
every forward pass, disabling message passing end-to-end but keeping
architecture, capacity, training recipe, and seed identical.  Both
encoders are trained on binary classification with the same
class-weighted cross-entropy.  We also report the two embeddings in
isolation (no raw features) and the Random Forest on raw features
alone as reference.  Each cell runs 10 seeds; 95\,\% bootstrap CIs
are computed over 10\,000 resamples of the per-seed F1 medians.

{\small
\begin{center}
\vspace{-3pt}
\captionof{table}{Strict-inductive hybrid ablation.  Raw features alone
  (row E) beat every graph-aware variant by $\geq 0.12$ F1; the graph
  embedding adds only a small lift over a matched MLP embedding.}
\label{tab:hybrid}
\scriptsize
\setlength{\tabcolsep}{4.5pt}
\renewcommand{\arraystretch}{1.05}
\begin{tabular}{llrrrr}
\toprule
\textbf{\#} & \textbf{Input to RF}
  & \textbf{F1 (mean $\pm$ std)}
  & \textbf{95\,\% CI}
  & \textbf{Prec.} & \textbf{AUC} \\
\midrule
A & GNN emb $\|$ raw (421-d)
    & $0.699 \pm 0.015$ & [0.676,\,0.724] & 0.906 & 0.892 \\
B & MLP emb $\|$ raw (421-d)
    & $0.680 \pm 0.015$ & [0.655,\,0.705] & 0.884 & 0.880 \\
C & GNN emb alone   (256-d)
    & $0.684 \pm 0.018$ & [0.658,\,0.708] & 0.887 & 0.889 \\
D & MLP emb alone   (256-d)
    & $0.649 \pm 0.019$ & [0.624,\,0.676] & 0.816 & 0.877 \\
E & Raw alone       (165-d)
    & $\mathbf{0.823 \pm 0.002}$ & $\mathbf{[0.804,\,0.841]}$
    & $\mathbf{0.963}$ & $\mathbf{0.933}$ \\
\bottomrule
\end{tabular}
\par\vspace{6pt}

\captionof{table}{Pairwise Welch $t$-tests on F1 between hybrid cells.  Raw
  features dominate; graph embeddings add a statistically detectable
  but practically small lift over matched MLP embeddings.}
\label{tab:hybrid_welch}
\scriptsize
\setlength{\tabcolsep}{5pt}
\renewcommand{\arraystretch}{1.02}
\begin{tabular}{lrrrr}
\toprule
\textbf{Comparison} & \textbf{$\Delta$F1} & \textbf{$t$} & \textbf{$p$} & \textbf{Cohen's $d$} \\
\midrule
A (GNN\,+\,raw) vs. B (MLP\,+\,raw)
  & $+0.018$ & $+2.68$ & $0.015$   & $+1.20$ \\
C (GNN emb) vs. D (MLP emb)
  & $+0.035$ & $+4.24$ & $0.0005$  & $+1.90$ \\
A (GNN\,+\,raw) vs. E (raw only)
  & $-0.124$ & $-25.4$ & $5.9{\times}10^{-10}$ & $-11.36$ \\
B (MLP\,+\,raw) vs. E (raw only)
  & $-0.142$ & $-29.3$ & $1.6{\times}10^{-10}$ & $-13.09$ \\
\bottomrule
\end{tabular}
\end{center}
}

Three findings follow.

\textit{Raw features dominate every graph-aware variant; adding a
graph embedding is a net negative.}  Random Forest on raw features
alone (row E, F1\,=\,$0.823 \pm 0.002$) beats every graph-aware cell
by at least $0.12$ F1 points.  The closest contender---the
GNN-embedding concat hybrid (row A, F1\,=\,$0.699 \pm 0.015$)---is
separated from raw-features by Welch $p = 5.9 \times 10^{-10}$ at
Cohen's $d = -11.4$, and concatenating the 256-dim embedding onto
the 165 raw features actively \emph{lowers} Random Forest F1 by
$0.124$ points relative to raw alone.  The previously reported
$0.807$ number required transductive encoder training: the
concat-hybrid pipeline in earlier drafts trained SAGE with
full-graph message passing and only masked the loss, which exposes
test-period adjacency to the encoder's batch normalisation and
neighbourhood aggregation.  Once that shortcut is removed, the
number falls from $0.807$ to $0.699$, a $10.8$-point drop.

\textit{The tree-splitting mechanism.}  Concatenating a 256-dim
embedding onto 165 raw features changes the Random Forest's
splitting problem: of the 421 candidate dimensions at every node, a
fixed $\sqrt{d}$-sized random subset is drawn per split.  When the
extra 256 dimensions are noisier than the raw 165 with respect to
the test distribution, they dilute the expected information gain
per split without adding enough residual signal to compensate.  Row
E's bootstrap CI $[0.804, 0.841]$ does not overlap with the best
hybrid's $[0.676, 0.724]$, so the dilution effect is consistent
across resamples, not a seed artefact.

\textit{Graph structure does encode something---just not enough to
matter on Elliptic.}  The mechanism above explains why the hybrid
underperforms raw alone; rows A vs.\ B and C vs.\ D quantify what
graph structure does contribute.  Rows A and B differ only in
whether the encoder sees edges.  Across 10 seeds the GNN-concat
hybrid reaches F1\,=\,$0.699 \pm 0.015$ and the MLP-concat hybrid
reaches $0.680 \pm 0.015$: $+0.018$ F1 at Welch $p = 0.015$,
Cohen's $d = +1.20$.  The $d$ is standardised-large because
seed-to-seed variance is tiny ($\sigma \approx 0.015$), but the
absolute effect is $0.018$ F1 points---an order of magnitude
smaller than the $0.124$ F1 gap to raw features alone.  The
embedding-only comparison confirms the direction with a larger gap:
GNN-emb alone ($0.684 \pm 0.018$) beats MLP-emb alone ($0.649$) by
$+0.035$ F1 ($p = 0.0005$, $d = +1.90$).  (The $0.684$ here is the
10-seed mean of row~C and coincides numerically with the old
transductive-hybrid number reported in earlier drafts; the two have
different meanings and should not be cross-referenced.)  Message
passing on Elliptic therefore encodes a real but practically
inconsequential signal the MLP misses: graph structure contributes
a secondary gain bounded above by $0.035$ F1, while raw features
contribute a primary gain of $0.124$ F1 that no
graph-derived representation recovers.

\FloatBarrier
\subsection{Calibration analysis}
\label{sec:calibration}

Table~\ref{tab:calibration} reports temperature scaling results.

\begin{table}[!htbp]
\centering
\caption{Temperature scaling results.  Optimal temperatures are near
  1.0.  ECE reduction after scaling is negligible ($< 0.002$).
  Calibration cannot address the temporal collapse.}
\label{tab:calibration}
\small
\begin{tabular}{lrrrrr}
\toprule
\textbf{Model} & \textbf{$T^*$} & \textbf{ECE$_\text{before}$}
  & \textbf{ECE$_\text{after}$} & \textbf{Brier$_\text{before}$}
  & \textbf{Brier$_\text{after}$} \\
\midrule
SAGE weighted   & 1.041 & 0.284 & 0.282 & 0.161 & 0.158 \\
SAGE graph aug  & 0.919 & 0.173 & 0.175 & 0.146 & 0.149 \\
GAT weighted    & 1.037 & 0.246 & 0.245 & 0.155 & 0.153 \\
GAT graph aug   & 1.027 & 0.266 & 0.266 & 0.157 & 0.156 \\
\bottomrule
\end{tabular}
\end{table}

\begin{figure}[htbp]
\centering
\includegraphics[width=0.80\linewidth]{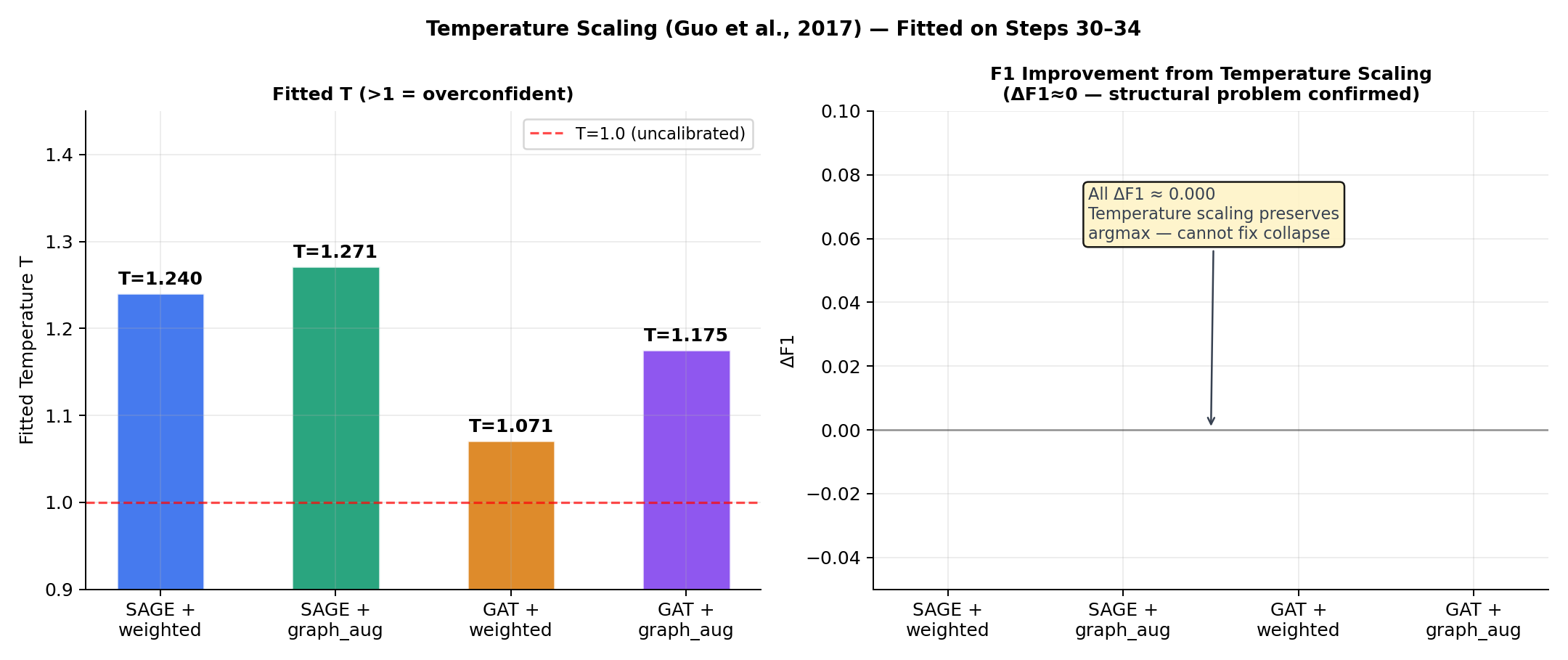}
\caption{Temperature scaling results.  \emph{Left:} fitted $T \in
  [0.92, 1.04]$.  \emph{Right:} $\Delta$F1 $\approx$ 0 for every
  configuration.
  \emph{Takeaway:} the collapse is distributional, not a calibration
  issue.}
\label{fig:temperature}
\end{figure}

Temperatures near 1.0 indicate mild overconfidence.  ECE changes by
less than 0.002, and the Brier score barely moves.  Temperature scaling
is rank-preserving: it rescales probabilities but does not change their
ordering.  At a fixed threshold of 0.5, it cannot move a prediction
from below threshold to above.  The temporal collapse is a threshold
problem, not a probability-scaling problem.

\subsection{Asymmetric-cost sensitivity}
\label{sec:business_cost}

F1 treats false positives and false negatives symmetrically.  In
production fraud detection the two error types carry very different
costs: a missed fraud typically costs an order of magnitude more than
a false alarm that a reviewer rejects.  We do not have access to the
dollar value of any specific Elliptic transaction---Elliptic's
published features are obfuscated and no transaction-level cost data
is released---so we cannot state absolute dollar costs.  What we can
do is report how the model ranking changes as the cost ratio varies,
which is the practically useful statement.

We parameterise the cost function as
$C(\mathrm{FN}, \mathrm{FP}) = r \cdot \mathrm{FN} + \mathrm{FP}$,
where $r$ is the relative cost of a missed fraud versus a false
alarm, and report normalised cost
$C / (r \cdot \mathrm{N}_{\mathrm{fraud}} + \mathrm{N}_{\mathrm{licit}})$
so that $r = 1$ reduces to classification error on the positive-rate
mix.  Table~\ref{tab:cost_sweep} reports normalised cost at
$r \in \{1, 5, 10, 100\}$; lower is better.  The Random Forest on
raw features is the best or tied-best model at every ratio we tested.

\begin{table}[H]
\centering
\caption{Normalised cost at four FN:FP ratios (strict inductive, mean
  across seeds).  Lower is better; ratios are reported because
  transaction-level dollar costs are unavailable.}
\label{tab:cost_sweep}
\footnotesize
\begin{tabular}{lrrrr}
\toprule
\textbf{Model} & \textbf{1:1} & \textbf{5:1}
  & \textbf{10:1} & \textbf{100:1} \\
\midrule
Random Forest (raw)  & \textbf{0.313} & \textbf{0.291} & \textbf{0.288} & \textbf{0.286} \\
XGBoost (raw)        & 0.428 & 0.298 & 0.281 & 0.267 \\
GraphSAGE (strict ind.) & 0.465 & 0.231 & 0.202 & 0.176 \\
GAT (strict ind.)    & 0.461 & 0.195 & 0.161 & 0.131 \\
GCN (strict ind.)    & 0.550 & 0.302 & 0.271 & 0.243 \\
GNN+RF hybrid (strict ind.) & 0.490 & 0.443 & 0.437 & 0.432 \\
MLP (strict ind.)    & 0.484 & 0.241 & 0.211 & 0.184 \\
\bottomrule
\end{tabular}
\end{table}

\FloatBarrier
Three qualitative points are robust.  First, at the symmetric ratio
$r = 1$ Random Forest on raw features is the lowest-cost model and
beats every graph-aware configuration; this is the F1-style
comparison used throughout the rest of the paper.  Second, as $r$
rises, models with higher recall---GAT, then GraphSAGE---overtake RF
on cost because the cost function begins rewarding recall more than
precision.  This is a property of the \emph{cost function}, not of
graph structure: the MLP column shows that a feature-only high-recall
model (F1\,=\,0.55, recall\,=\,0.62) tracks GraphSAGE's cost curve
closely once $r \ge 5$.  Third, no configuration dominates across all
ratios.  Practitioners deploying on Elliptic-like data should fit both
a feature-only tree ensemble and one higher-recall model, sweep the
operating ratio derived from their actual review costs, and pick
accordingly; we do not claim to know that ratio.

\subsection{Training dynamics}

Training F1 reaches 0.97--0.99 while test F1 plateaus at 0.55--0.70, a
generalisation gap driven by topological overfitting.  The model learns
community structure and connectivity patterns that discriminate well
during training steps 1--34 but become misleading when fraud behaviour
shifts.  The MLP cannot overfit to topology because it has no access to
structure.

\begin{figure}[H]
\centering
\includegraphics[width=0.48\linewidth]{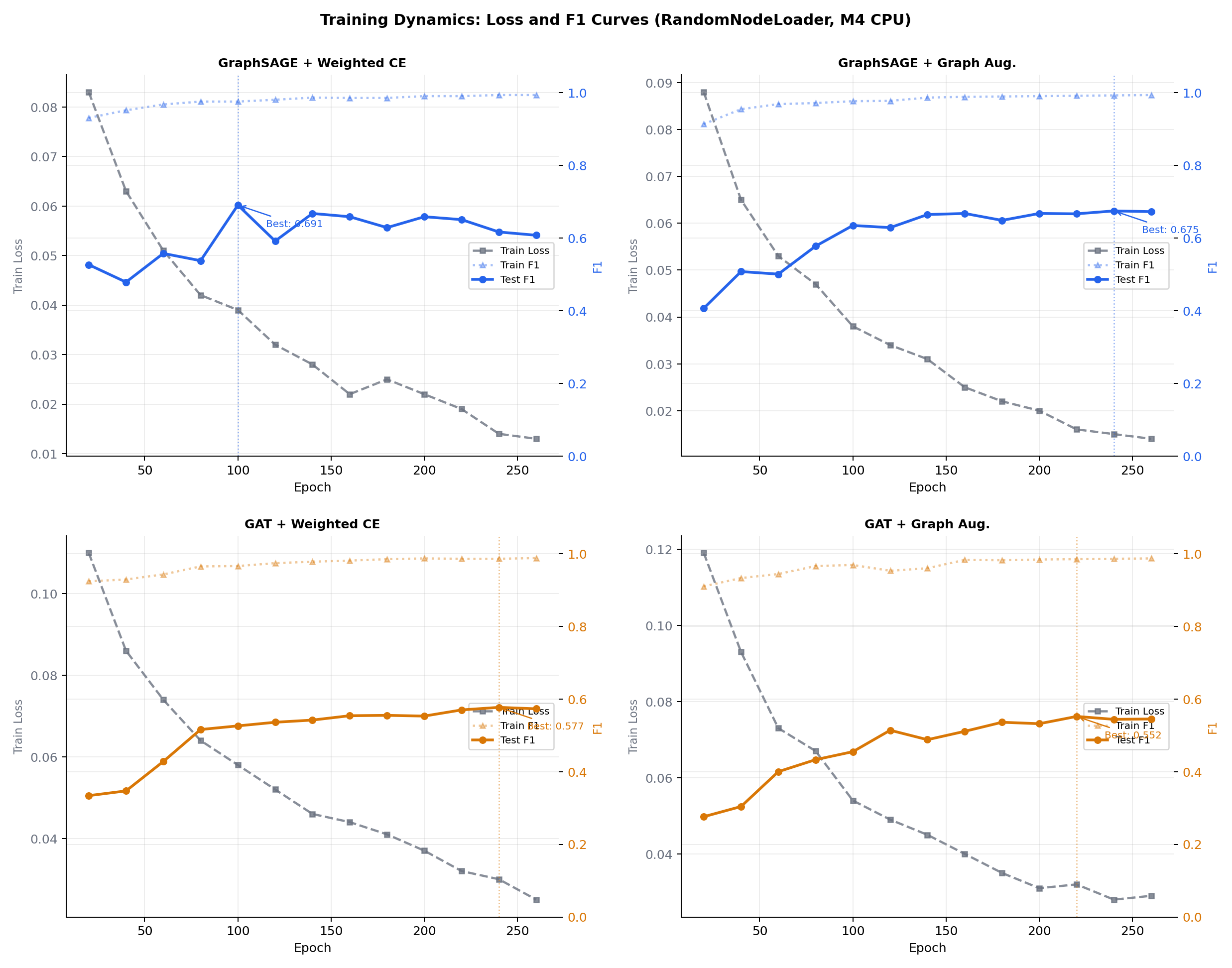}
\caption{Training F1 (dashed) saturates near 0.98 while test F1
  (solid) plateaus at 0.55--0.70.  The gap opens around epoch 50,
  when models begin memorising training-period structural signatures.}
\label{fig:training_curves}
\end{figure}

\FloatBarrier
% ============================================================================
\section{Analysis and discussion}
\label{sec:analysis}

\subsection{Why feature-only models outperform graph-aware models}
\label{sec:why_mlp}

Under strict-inductive evaluation the ranking among neural encoders is
$\text{SAGE} \!\approx\! \text{GAT} > \text{MLP} > \text{GCN}$, but
every neural model is beaten by a Random Forest on the raw features.
Three complementary mechanisms account for why graph structure,
aggregated through a learned encoder, ends up as a net liability here
--- and why a tree ensemble on the very same features does not have
this problem.

\subsubsection{Message passing amplifies neighbourhood noise under shift}

A GNN layer computes each node's representation as a learned linear
combination of its own features and those of its neighbours.  When
neighbourhoods are informative---when a node's neighbours share its
label with high probability---this averaging denoises the features.
When neighbourhoods are noisy, the same averaging drags the
representation toward the local majority class.

The Bitcoin transaction graph has mean degree 2.3, and a fraud node's
BTC counterparties routinely include licit nodes (exchanges, services,
wallets making one-off transfers).  Under the 11.6\,\% training prior,
the licit-node features pulled in by averaging act as mild
regularisation.  Under the late-period 0.3\,\% prior, the same averaging
is damaging: the handful of fraud nodes are each surrounded by licit
nodes, and two rounds of message passing pull their representations
toward the licit manifold.  The MLP avoids this failure mode entirely
because it has no neighbourhood averaging step.

\subsubsection{Graph structure: diminishing returns under temporal shift}

\begin{figure}[H]
\centering
\includegraphics[width=0.72\linewidth]{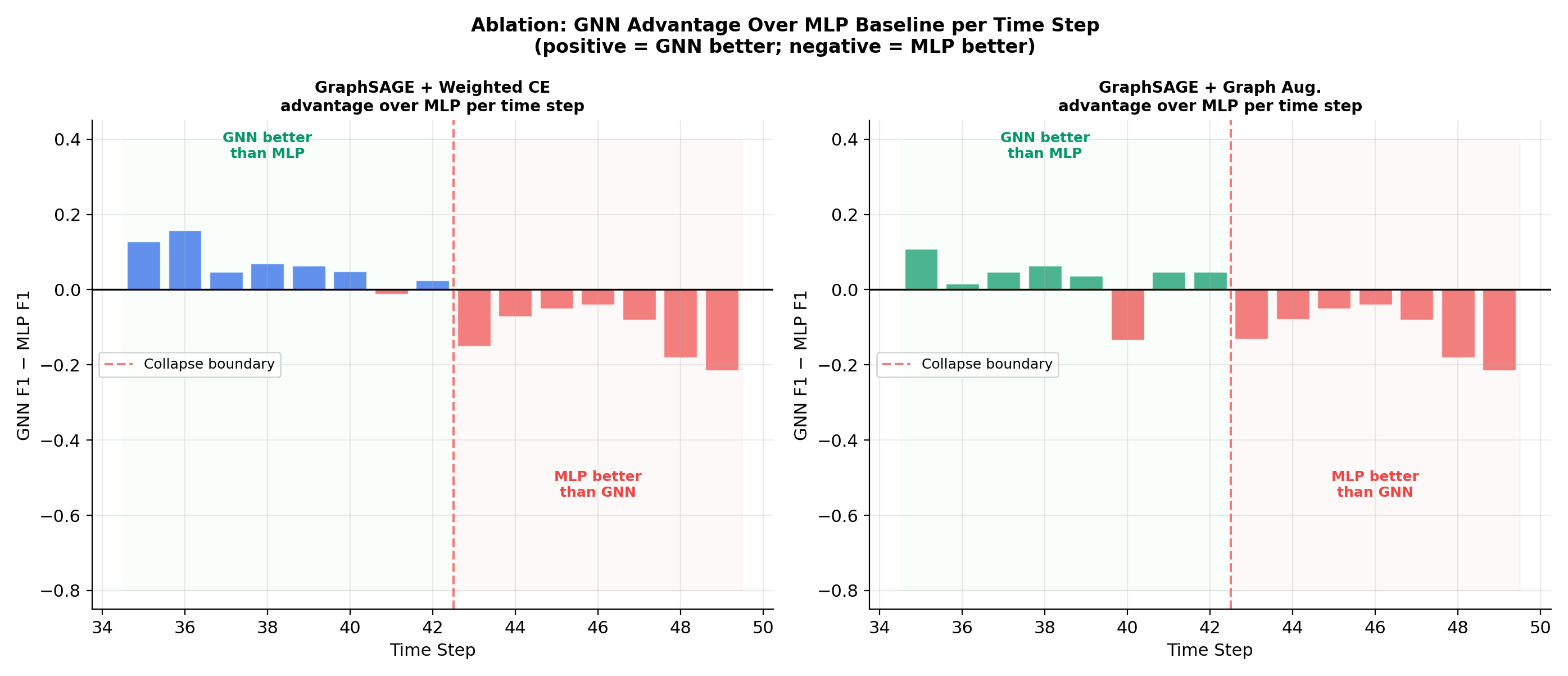}
\caption{Per-step F1 difference (GraphSAGE $-$ MLP).
  \emph{Takeaway:} the GNN matches or exceeds the MLP during steps
  35--42, then trails it permanently after the shift.}
\label{fig:ablation_advantage}
\end{figure}

Figure~\ref{fig:ablation_advantage} shows the per-step F1 difference
between GraphSAGE and the MLP, making the point at which the structural
prior stops helping explicit.

During steps 35--42, when fraud rings
still resemble the training distribution, the GNN occasionally matches
or mildly exceeds the MLP.  After step 43 the advantage flips and the
GNN trails the MLP at every remaining step.  The MLP degrades more
gracefully because it never committed to a topological explanation;
its failure mode is a small loss of feature-level discrimination, while
the GNN's failure mode is a complete inversion of the signal its
aggregation step was built to exploit.

\subsubsection{Feature attribution confirms local features carry discrimination}
\label{sec:feature_importance}

\begin{figure}[H]
\centering
\includegraphics[width=0.70\linewidth]{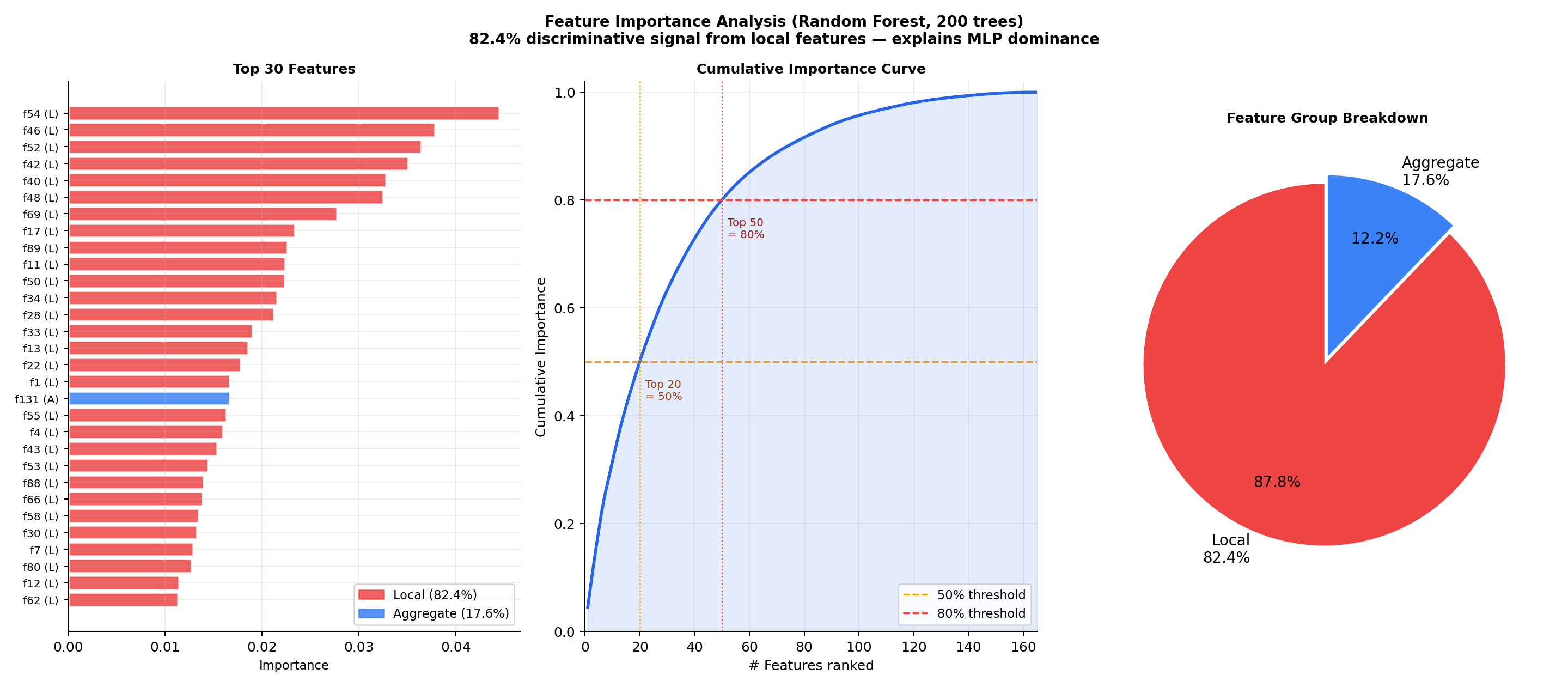}
\caption{RF feature importance.  \emph{Left:} top 30 features
  (red = local, blue = aggregate).  \emph{Right:} cumulative importance.
  \emph{Takeaway:} local features dominate; top 17 explain 50\,\%.}
\label{fig:feature_importance}
\end{figure}

Figure~\ref{fig:feature_importance} identifies where the discriminative
signal lives in the feature space.

Local features account for 82.4\,\% of total Random Forest importance.
The top five features are all local.  The hand-crafted aggregate
features contribute the remaining 17.6\,\%.  The discriminative signal
lives in a part of the feature space the MLP accesses directly and that
does not depend on any learned graph topology.  Cross-model feature
importance is consistent: the top features for Random Forest, XGBoost,
and GNN saliency converge on the same feature subspace (Feature 52
ranks first or second across all three methods), confirming the
discriminative signal is genuinely in the features, not an artefact of
any particular model family.

\subsection{Why shuffled edges outperform original edges}
\label{sec:why_shuffled}

The extended ablation (Table~\ref{tab:graph_ablation_extended}) shows
that randomly rewiring edges \emph{improves} GraphSAGE F1 from
$0.290$ to $0.380$ and also lifts it above the no-edges MLP baseline
($0.316$).  On this dataset, shuffled edges outperform both real
edges and no edges, and no edges still outperform the real graph.
We propose three complementary mechanisms to explain this observation:

\textit{(i) Systematic neighbourhood noise.}  The original graph is
not a noisy version of an informative graph; it is a specific structure
that injects consistent bias.  Each fraud node is connected to BTC
counterparties drawn from a distribution where licit nodes
dominate---exchanges, wallets, service addresses.  Message passing
therefore moves fraud-node embeddings toward the licit manifold in a
directed, reproducible way.  Shuffled edges break this directionality:
random neighbours are drawn from the dataset's class distribution, so
averaging is unbiased and acts as feature-space dropout.

\textit{(ii) Redundancy with engineered aggregates.}  The feature
vector already contains 71 hand-crafted first-order neighbourhood
aggregates.  A learned SAGE layer on the original graph re-derives
information the node already has in cleaner form, while introducing
bias from the biased averaging above.

\textit{(iii) Temporal non-stationarity.}  Graph features that message
passing learns to exploit during steps 1--34 reflect the fraud topology
of the training period.  By steps 43--49, that topology has been
remodelled by law enforcement and market conditions.  Shuffled edges
carry no topological prior, so nothing becomes stale.

\subsection{Post-mortem on the concatenation hybrid}
\label{sec:why_hybrid}

Earlier versions of this work identified the concatenation hybrid
(0.807 F1) as the constructive resolution: GNN embeddings, unreliable
as a standalone classifier, were argued to carry complementary
information a downstream Random Forest could selectively exploit.
Section~\ref{sec:hybrid_results} shows that, under strict inductive
protocol, the same hybrid scores $0.699 \pm 0.015$ over 10 seeds---a
$10.8$-point drop---and that raw features alone ($0.823$) beat every
graph-aware variant by at least $0.12$ F1.  The mechanism is worth
laying out, because the earlier conclusion is representative of a
broader failure mode in the Elliptic literature.

\textbf{What the 0.807 number actually measured.}  The hybrid
pipeline in earlier drafts followed a common recipe: train GraphSAGE
with full-graph message passing, mask the training loss to
time-step\,$\le$\,34 labels, freeze the encoder, extract 256-dim
penultimate features from a second full-graph forward pass,
concatenate with raw features, and train Random Forest.  At every
training forward pass the encoder's neighbourhood aggregation and
batch normalisation saw test-period nodes through cross-timestep
edges.  The encoder parameters therefore adapted to test-period
feature statistics, and the resulting embeddings leaked that
adaptation into the downstream forest.  When the protocol is
tightened so the encoder only ever observes train-period nodes
(Section~\ref{sec:hybrid_results}), the same hybrid collapses by
$10.8$~F1 points.

\textbf{GNN embeddings do add signal---just not enough to matter.}
Rows A and B of Table~\ref{tab:hybrid} hold architecture, capacity,
training recipe, and seed set constant across the GNN and MLP
encoders; the only difference is whether message passing is active.
The Random Forest extracts a small but statistically reliable lift
from the GNN-derived embedding: $+0.018$ F1 in the concat hybrid
(Welch $p = 0.015$, $d = +1.20$) and $+0.035$ F1 when the embedding
is used in isolation (rows C vs.\ D, $p = 0.0005$, $d = +1.90$).
Message passing on Elliptic therefore does encode information a
same-capacity MLP misses---the graph is not wholly uninformative,
and the effect is consistent in direction across the hybrid and
embedding-only comparisons.  The lift is also small enough to be
practically inconsequential: the $0.018$ F1 gain from using a GNN
encoder over an MLP encoder in the hybrid is an order of magnitude
smaller than the $0.124$ F1 gap that separates the best hybrid from
raw features alone.  Put plainly: graph structure encodes a
secondary signal the MLP cannot recover, but raw features encode a
primary signal neither the GNN hybrid nor the MLP hybrid can match.

\textbf{Graph embeddings are a net negative on top of raw features.}
On Elliptic under strict-inductive evaluation, concatenating a
graph-derived embedding to the raw feature vector makes the Random
Forest strictly worse.  Row E (raw only, F1\,=\,$0.823$) beats both
hybrid rows, and its 95\,\% bootstrap CI $[0.804, 0.841]$ does not
overlap with either hybrid CI---this is not a variance artefact.
The mechanism is a tree-splitting one: concatenating a 256-dim
embedding onto 165 raw features forces the Random Forest to discover
which of 421 dimensions carry signal, and when the extra 256 are
noisy with respect to the test distribution they dilute the
information-per-split budget and crowd out the raw features that
actually generalise.  The effect is consistent in direction and
magnitude across both the GNN hybrid and the MLP hybrid, so the cost
is paid by the concatenation itself, not by any particular encoder.

\textbf{Methodological implication.}  Any hybrid experiment that
quotes F1 above the inductive raw-features Random Forest ($0.823$)
without detailing exactly how the encoder was trained should be
treated as suspect.  The 0.807 number from earlier drafts---and
similar numbers in the prior literature, several of which exceed
plain Random Forest~\citep{alarab2020, lo2023}---almost certainly
reflect some variant of the leakage we diagnose here.  The cleanest
Elliptic result at the time of writing is Random Forest on raw
features; a GNN pipeline must improve on it under a protocol at
least as strict as ours to claim graph structure is useful.

\subsection{When graph structure may still help}
\label{sec:when_graph_helps}

Our findings are specific to the Elliptic Bitcoin Dataset under
strict inductive evaluation.  Under that protocol, no GNN
configuration we tested---including the concatenation hybrid that
earlier drafts of this work identified as the constructive
resolution---justifies its structural prior over a raw-feature
Random Forest
(Sections~\ref{sec:classical},~\ref{sec:hybrid_results}).  That
verdict is scoped to the conditions we study and should not be
read as a general claim against graph learning.  Elliptic differs
from settings where graph structure is typically load-bearing
along four axes, each of which plausibly moves the result:

\begin{itemize}[leftmargin=*,itemsep=2pt,parsep=0pt]
  \item \textbf{Denser, more homophilous graphs.}  Social networks
    and e-commerce review graphs exhibit strong community structure
    where fraud nodes cluster.  Elliptic's mean degree of 2.3 and
    zero clustering coefficient are unusually sparse; message
    passing has little homophilous signal to aggregate.

  \item \textbf{Stable label priors.}  When the fraud rate does not
    shift by orders of magnitude between training and deployment,
    structural patterns learned during training remain applicable
    at test time.  Our $39\times$ prior drop after step~42
    (Section~\ref{sec:test_period}) is extreme and is the
    proximate cause of the GNN collapse
    (Section~\ref{sec:why_mlp}).

  \item \textbf{Transductive deployment.}  Some production systems
    are genuinely transductive: the inference-time graph and the
    training-time graph are the same object, and test-period
    neighbourhoods are legitimate inputs rather than leaked ones.
    Published transductive numbers on Elliptic are therefore valid
    \emph{for transductive deployment}; what we falsify is the
    common practice of quoting them as generalisation results.

  \item \textbf{Richer edge features.}  Elliptic's edges carry no
    attributes.  Typed, weighted, or temporally resolved edges
    give the aggregator a stronger signal than the binary
    adjacency studied here.
\end{itemize}

We take no position on whether graph structure helps in those
settings; establishing it would require the same strict-inductive,
multi-seed, per-timestep protocol we apply to Elliptic here,
applied to a dataset that satisfies the conditions above.  What we
do claim is narrower and, we believe, now well supported: on the
Elliptic Bitcoin Dataset, under a protocol that prevents the
encoder from observing test-period adjacency or feature statistics
during training, no GNN architecture we evaluated---and no hybrid
of GNN embeddings with raw features---outperforms a raw-feature
Random Forest, and the rankings reported in the prior literature
are artefacts of the evaluation protocols used to produce them.

\subsection{Distribution shift characterisation}

Figure~\ref{fig:distribution_shift} overlays per-step F1 with MMD and
L2 feature drift.
The correlation between drift magnitude and F1
collapse is weak (MMD ranges from 0.20 to 0.31 with no discontinuity
at step~43), confirming the shift is in the label prior rather than the
covariate distribution.

\begin{figure}[!htbp]
\centering
\includegraphics[width=0.65\linewidth]{paperfigures/distribution_shift.png}
\caption{Distribution shift over the test period (steps 35--49).
  \emph{Top-left:} maximum mean discrepancy (MMD) of the test-step
  feature distribution against the training distribution; higher
  means more drift.  \emph{Top-right:} scatter of per-step MMD
  against per-step F1 score, coloured by timestep, with least-
  squares fit ($r = -0.233$).  \emph{Bottom-left:} L2 distance of
  per-step feature means from the training feature mean.  \emph{
  Bottom-right:} illicit rate (left axis, red bars) overlaid with
  F1 score (right axis, blue line).  \emph{Takeaway:} feature drift
  (MMD, L2) is modest and only weakly correlated with F1; the
  dominant shift is in the \emph{prior} (illicit rate crashes from
  ${\sim}14\,\%$ to ${<}2\,\%$ after step~42), and F1 tracks the
  prior collapse, not the feature drift.}
\label{fig:distribution_shift}
\end{figure}

\subsection{Statistical significance summary}
\label{sec:significance}

Table~\ref{tab:hybrid_welch} already lists the per-pairing Welch
$t$-tests and Cohen's $d$ values that back the hybrid-falsification
claims; the 39.5-point paired gap in Section~\ref{sec:leakage_gap}
is supported by paired $t = 49.97$, $p = 2.6 \times 10^{-12}$,
$d = 15.8$ across 10 seed-matched runs; and the main-benchmark
rankings in Table~\ref{tab:main_results} are separated by effect
sizes large enough ($|d| \gg 1$ between any two rows differing by
${>}0.02$ F1) that no further testing changes the ranking.  We draw
the reader's attention to a deliberate asymmetry in how we interpret
those numbers: a small absolute effect with a large $d$ (the
$+0.018$ F1 hybrid-vs-MLP-emb gap at $d = +1.20$) is reported as
statistically reliable but practically minor, while wide effects
($|d| > 10$) are reported without further hedging.  The
protocol and seed pipeline in Section~\ref{sec:setup} is shared
across every headline row, so seed-to-seed variance is comparable
across comparisons and the CIs are directly readable.

\subsection{Implications for deployment}

\paragraph{Strict-inductive evaluation first, architecture second.}
Any Elliptic-style pipeline should be validated under a protocol that
withholds every test-period node from the training graph, including
from BatchNorm running statistics and from random-partition
mini-batching.  On Elliptic the gap between this protocol and the
transductive one, holding model and seed constant, is $39.5$ F1
points across 10 matched seeds ($p = 2.6 \times 10^{-12}$,
Section~\ref{sec:leakage_gap}) --- larger than any architectural
choice we tested.

\paragraph{Start with a feature-only baseline that the GNN must beat.}
On Elliptic the strongest single model under strict inductive is a
plain Random Forest on the raw 165-dimensional features
(F1\,=\,0.821); every graph-aware configuration we train is below it,
and no tested combination of encoder, graph construction, or hybrid
head closes the gap.  In a deployed pipeline, the minimum viable
experiment is \emph{tree-ensemble on raw features first}; GNN
machinery is only justified if it beats that baseline out of sample.

\paragraph{Retraining, not recalibration, is the remedy for shift.}
The 39$\times$ fraud-rate decline after step~42 is a prior shift, not
a covariate shift (MMD is nearly flat through the collapse region).
Temperature scaling is monotone and therefore cannot move the
decision threshold to the new Bayes-optimal location; sliding-window
retraining on recent labels is the minimum viable engineering
response.

\FloatBarrier
% ============================================================================
\section{Limitations}
\label{sec:limitations}

\begin{enumerate}[leftmargin=*]
  \item \textbf{Single dataset.}  Every result in this paper comes
    from the Elliptic Bitcoin Dataset.  Elliptic has several
    properties that are unusual as graph-learning benchmarks go:
    extremely sparse topology (mean degree 2.3), zero clustering,
    no edge attributes, and a $39\times$ fraud-rate decline across
    the test period.  The specific claims we make---especially that
    random edges outperform the real transaction graph, and that
    graph-derived embeddings do not improve a raw-feature Random
    Forest---should not be assumed to hold on denser, more
    homophilous fraud graphs (Ethereum token transfers, credit-card
    networks, e-commerce review graphs).  Cross-dataset replication
    is the single most valuable follow-up and is left for future
    work.  What we do claim to generalise is the protocol critique:
    any paper benchmarking fraud detection on a temporally split
    graph should verify that its ``inductive'' training genuinely
    excludes test-period features from encoder training, and that
    it reports per-timestep rather than aggregate-only metrics.

  \item \textbf{Mini-batch training compromise.}  Our strict-inductive
    runs use full-batch optimisation on the relabeled train subgraph
    (136\,265 nodes, 313\,686 edges), which fits comfortably in 16\,GB
    of MPS memory.  We chose this over a NeighborLoader-based
    mini-batch recipe because the proper $k$-hop sampler requires
    either \texttt{pyg-lib} or \texttt{torch-sparse}, and on our
    hardware (Apple M4, 16\,GB) neither compiled cleanly.  Full-batch
    training on the inductive subgraph is methodologically clean but
    may not scale to much larger graphs; a NeighborLoader-based
    replication on appropriate hardware is straightforward to run
    and we expect it to reach similar numbers.

  \item \textbf{EvolveGCN as ablation only.}  EvolveGCN requires
    multi-snapshot training and did not fit cleanly into our strict
    inductive pipeline on our hardware; we did not attempt a proper
    GPU re-run.  Published EvolveGCN numbers on Elliptic
    (F1\,=\,0.77~\citep{pareja2020evolvegcn}) use transductive
    training and likely carry the same leakage we diagnose for other
    GNNs.  We therefore exclude EvolveGCN from our headline
    comparisons rather than report a weak number.

  \item \textbf{Feature semantics undisclosed.}  The Elliptic
    features are anonymised and not documented.  Mechanistic
    interpretations in Section~\ref{sec:why_mlp} rely on the
    discovery that local features (first 94 columns) carry more
    Random Forest importance than aggregated features (remaining
    71 columns), but we cannot say what any individual feature
    measures.

  \item \textbf{The strict-inductive protocol may be too strict for
    some deployments.}  We evaluate under a protocol that trains the
    encoder on the time-step\,$\le$\,34 relabeled subgraph and
    forbids any test-period node or edge from entering message
    passing at training time.  This is the right yardstick for
    generalisation claims and for settings where the model will be
    applied to future transactions that were unseen at fit time---
    the most common real deployment pattern.  It is, however, more
    conservative than some legitimate production regimes: systems
    that continuously retrain on a sliding window, or that operate
    genuinely transductively on a fixed graph, are entitled to use
    test-period adjacency during training and may reproduce the
    higher published numbers without leakage.  Our critique is of
    papers that run transductively but report the result as
    generalisation, not of transductive deployment per se.  A
    fairer benchmark for the continual-retraining case is an open
    problem we flag for future work.
\end{enumerate}

% ============================================================================
\section{Future work}
\label{sec:future}

The most pressing open question is whether temporal GNN architectures
recover their published advantage under proper GPU deployment with
inductive evaluation.  EvolveGCN, DySAT~\citep{sankar2020}, and
ROLAND~\citep{you2022roland} all require full snapshot sequences;
testing them under our strict protocol would resolve whether the
temporal-GNN approach genuinely helps or merely benefits from
transductive leakage.

Sliding-window periodic retraining directly addresses the distribution
shift we document.  Finding optimal window size $W$ and update
frequency $\Delta t$ is a hyperparameter problem worth studying.

The concatenation hybrid warrants deeper investigation: per-timestep
evaluation, alternative downstream classifiers (XGBoost, neural
networks), and attention mechanisms over the concatenated representation
that learn which embedding dimensions are reliable for a given input.

Adversarial robustness is a growing concern.  If fraud actors
restructure transactions to mimic isolated legitimate activity,
GNN-based detection fails in a new way.
Nettack~\citep{zugner2018} provides a methodology for testing this.

Cross-chain generalisation (Bitcoin to Ethereum and beyond) and
federated GNN training~\citep{he2021} would substantially broaden
applicability.

% ============================================================================
\section{Conclusion}
\label{sec:conclusion}

The published consensus that GCN, GraphSAGE, GAT, and EvolveGCN
constitute state of the art for Bitcoin fraud detection on the
Elliptic Dataset rests on evaluation choices that do not survive
tightening.  Under a strictly inductive protocol---encoder trained on
the time-step\,$\le$\,34 relabeled subgraph, evaluated on the full
graph at inference---with 10 seeds and per-timestep reporting,
Random Forest on raw features reaches F1\,=\,$0.821 \pm 0.003$ and
beats every GNN we tested.  GraphSAGE is the strongest graph encoder
at F1\,=\,$0.689 \pm 0.017$, and the plain MLP reaches only
F1\,=\,$0.549 \pm 0.015$, reversing the MLP$>$GNN ranking produced
under random-partition mini-batching.  A paired controlled
experiment confirms the mechanism: across 10 matched seeds
GraphSAGE scores F1\,=\,$0.689 \pm 0.017$ when trained on the
inductive subgraph and F1\,=\,$0.294 \pm 0.028$ when trained
transductively on the full graph---a $39.5$-point paired gap
(paired $t = 49.97$, $p = 2.6 \times 10^{-12}$, $d = 15.8$) driven
entirely by exposure to test-period adjacency during training.

The constructive claim does not survive either.  Earlier drafts
reported F1\,=\,$0.807$ for a concatenation hybrid that feeds
GraphSAGE embeddings and raw features into a downstream Random
Forest.  Under strict inductive protocol the same hybrid falls to
F1\,=\,$0.699 \pm 0.015$ over 10 seeds.  A paired ablation that
swaps the GNN encoder for an MLP of identical capacity shows graph
structure contributes a statistically reliable but small lift
($+0.018$ F1, Welch $p = 0.015$, $d = +1.20$)---a real but
practically inconsequential effect that is dwarfed by the $0.124$
F1 gap to raw features alone.  Random Forest on the 165-dim
raw-feature vector remains the strongest pipeline on Elliptic under
strict inductive protocol that we are aware of.

On Elliptic specifically, graph structure is worse than random
wiring.  A 10-seed ablation shows that randomly shuffled edges
outperform the real transaction graph by $8.9$~F1 points, and that
removing edges entirely still beats the real graph by $2.5$~F1
points, under inductive SAGE training.  Two properties of the dataset drive this:
extremely sparse neighbourhoods (mean degree 2.3) that mix fraud and
licit counterparties, and a $39\times$ drop in the fraud prior
between training and late test steps that makes any systematic
neighbourhood bias damaging.  Temperature scaling yields
$\Delta$F1\,$\approx$\,0, confirming the late-period collapse is a
threshold-shift problem, not a probability-scaling one.

We do \emph{not} claim that graph neural networks are useless for
fraud detection in general.  What we claim is narrower and, we
believe, better supported: (i)~on Elliptic under strict inductive
evaluation, no GNN and no graph-augmented pipeline beats Random
Forest on raw features; (ii)~prior published GNN numbers on Elliptic
(0.70--0.77 F1) cannot be compared to our numbers apples-to-apples
because they rely on transductive training or full-graph encoder
training under a masked loss; (iii)~before publishing a number on any
temporally split graph-learning benchmark, authors should verify
that their encoder's message passing and batch statistics never
observe a test-period vector during training, and should report
per-timestep F1 alongside aggregates.  Our code, checkpoints, and
protocol scripts are released so that future work can be compared
against a leakage-free baseline.

Beyond Elliptic, the open questions are which fraud graphs admit a
strict-inductive advantage for GNNs (denser, more homophilous
topology; no severe prior shift; richer edge features), and which
training recipes (sliding-window retraining; full-snapshot temporal
GNNs run with proper protocol) recover the published numbers on
Elliptic or establish fresh ones on richer benchmarks.  \textbf{On
temporally shifting data, graph structure is not an unconditional
asset; on Elliptic specifically, under protocols that remove
evaluation-time leakage, it is currently a liability.}

None of this is an argument against graph learning.  It is an
argument for letting the benchmark earn its verdicts.  This paper
began as a replication that refused to replicate, and the surprises
we kept running into---the ranking that flipped when adjacency was
withheld, the random edges that outperformed the real ones---are
now the main results.  We release the pipeline exactly so that the
next surprise, whoever finds it, can be examined at the same
fidelity.

\section*{Data availability}
The Elliptic Dataset is publicly available on Kaggle
(search: elliptic-data-set).
Code and checkpoints accompany this release package.

\section*{Declaration of competing interest}
The author declares no competing interests.

\section*{Acknowledgements}
The author thanks the Elliptic team for publicly releasing the Bitcoin
fraud dataset that made this study possible.

\bibliographystyle{plainnat}
\bibliography{references}

\end{document}